\documentclass[10pt,twocolumn,letterpaper]{article}

\usepackage[pagenumbers]{cvpr} % To force page numbers, e.g. for an arXiv version

\usepackage[accsupp]{axessibility}  % Improves PDF readability for those with disabilities.
\usepackage{graphicx}
\usepackage{amsmath}
\usepackage{amssymb}
\usepackage{booktabs}
\usepackage{bm}
\usepackage{xcolor}
\usepackage{color, colortbl}
\usepackage{algorithmic}
\usepackage[linesnumbered,ruled,vlined]{algorithm2e}
\usepackage{threeparttable}
\usepackage[pagebackref,breaklinks,colorlinks]{hyperref}
\usepackage[capitalize]{cleveref}
\usepackage{float}
\usepackage{bbding}
\usepackage{multirow}

\makeatletter
\renewcommand\paragraph{
  \@startsection{paragraph} % name
  {4} % level
  {\z@} % indent
  {.5em \@plus1ex \@minus.2ex} % beforeskip
  {-1.5em} % afterskip
  {\normalfont\normalsize\bfseries} % style
}
\makeatother

\makeatletter
\def\@fnsymbol#1{\ensuremath{\ifcase#1\or \textsuperscript{~\Envelope}\or \ddagger\or
   \mathsection\or \mathparagraph\or \|\or **\or \dagger\dagger
   \or \ddagger\ddagger \else\@ctrerr\fi}}
\makeatother

\crefname{section}{Sec.}{Secs.}
\Crefname{section}{Section}{Sections}
\Crefname{table}{Table}{Tables}
\crefname{table}{Tab.}{Tabs.}

\def\cvprPaperID{6097} % *** Enter the CVPR Paper ID here
\def\confName{CVPR}
\def\confYear{2022}

\newcommand{\method}{RIPU\xspace}
\newcommand{\region}{RA\xspace}
\newcommand{\pixel}{PA\xspace}
\newcommand{\Neighbor}{\mathcal{N}}
\newcommand{\KSquareNeighbor}{$k$\textit{-square-neighbors}\xspace}
\newcommand{\Purity}{\mathcal{P}}
\newcommand{\Uncertainty}{\mathcal{U}}
\newcommand{\Entropy}{\mathcal{H}}
\newcommand{\Probability}{\mathbf{P}}

\newcommand{\Losssup}{\mathcal{L}_{sup}}
\newcommand{\Lossce}{\mathcal{L}_{CE}}
\newcommand{\Lossneg}{\mathcal{L}_{nl}}
\newcommand{\Losssoft}{\mathcal{L}_{cr}}
\newcommand{\Label}{\mathbf{Y}}
\newcommand{\Pseudo}{\widehat{\mathbf{Y}}}
\newcommand{\ActiveLabel}{\widetilde{\mathbf{Y}}}
\newcommand{\Image}{\mathbf{I}}
\newcommand{\ActiveScore}{\mathcal{A}}

\newcommand{\Output}{\mathbf{P}}
\newcommand{\argmax}{\mathop{\arg\max}}

\newcommand{\sumc}{\sum_{c = 1}^C}
\newcommand{\sums}{\sum_{(i, j)\in \Image}}
\newcommand{\sumss}{\sum_{(i, j)\in \Image_s}}
\newcommand{\sumst}{\sum_{(i, j)\in \Image_t}}
\renewcommand{\paragraph}[1]{\vspace{1.25mm}\noindent\textbf{#1}}
\newcommand{\MY}[1]{{\textcolor{red}}}  % comments

\definecolor{Gray}{gray}{0.9}

\begin{document}

\title{Towards Fewer Annotations: Active Learning via Region Impurity and Prediction Uncertainty for Domain Adaptive Semantic Segmentation}

\author{Binhui Xie$^{1}$ \quad Longhui Yuan$^{1}$ \quad Shuang Li$^{1}$\thanks{Corresponding author} \quad Chi Harold Liu$^{1}$ \quad Xinjing Cheng$^{2,3}$ \\
$^{1}$School of Computer Science and Technology, Beijing Institute of Technology \\
$^{2}$School of Software, BNRist, Tsinghua University \quad $^{3}$Inceptio Technology \\
{\tt\small \{binhuixie,longhuiyuan,shuangli,chiliu\}@bit.edu.cn \quad cnorbot@gmail.com}
}
\maketitle
%%%%%%%%%%%%%%%%%%%%%%%%%%%%%%%%%%%%%%%%%%%%%%%%%%%%%%%%%%%%%%%%%%%%%%%%%%%%%%%%%%%%%%%%%%%%%%%%%%%

%%%%%%%%% ABSTRACT
\begin{abstract}
    Self-training has greatly facilitated domain adaptive semantic segmentation, which iteratively generates pseudo labels on unlabeled target data and retrains the network. However, realistic segmentation datasets are highly imbalanced, pseudo labels are typically biased to the majority classes and basically noisy, leading to an error-prone and suboptimal model. In this paper, we propose a simple region-based active learning approach for semantic segmentation under a domain shift, aiming to automatically query a small partition of image regions to be labeled while maximizing segmentation performance. Our algorithm, Region Impurity and Prediction Uncertainty (\method), introduces a new acquisition strategy characterizing the spatial adjacency of image regions along with the prediction confidence. We show that the proposed region-based selection strategy makes more efficient use of a limited budget than image-based or point-based counterparts. Further, we enforce local prediction consistency between a pixel and its nearest neighbors on a source image. Alongside, we develop a negative learning loss to make the features more discriminative. Extensive experiments demonstrate that our method only requires very few annotations to almost reach the supervised performance and substantially outperforms state-of-the-art methods. The code is available at \url{https://github.com/BIT-DA/RIPU}.
\end{abstract}

%%%%%%%%% INTRO
\section{Introduction}
\label{sec:intro}

%%%%%%%%%%%%%%%%%%%%%%%%%%%%%%%%%%%%%%%%%%%%%%%%%%%%%%%%%%%%%%%%%%%%%%%%%%%%%%%%%%%%%%%%%%%%%%%%%%%%%%%%%%%%%%%%%%%%%%%%%%%%%
\begin{figure}[!htbp]
    \centering
    \begin{subfigure}{0.22\textwidth}
        \includegraphics[width=\linewidth]{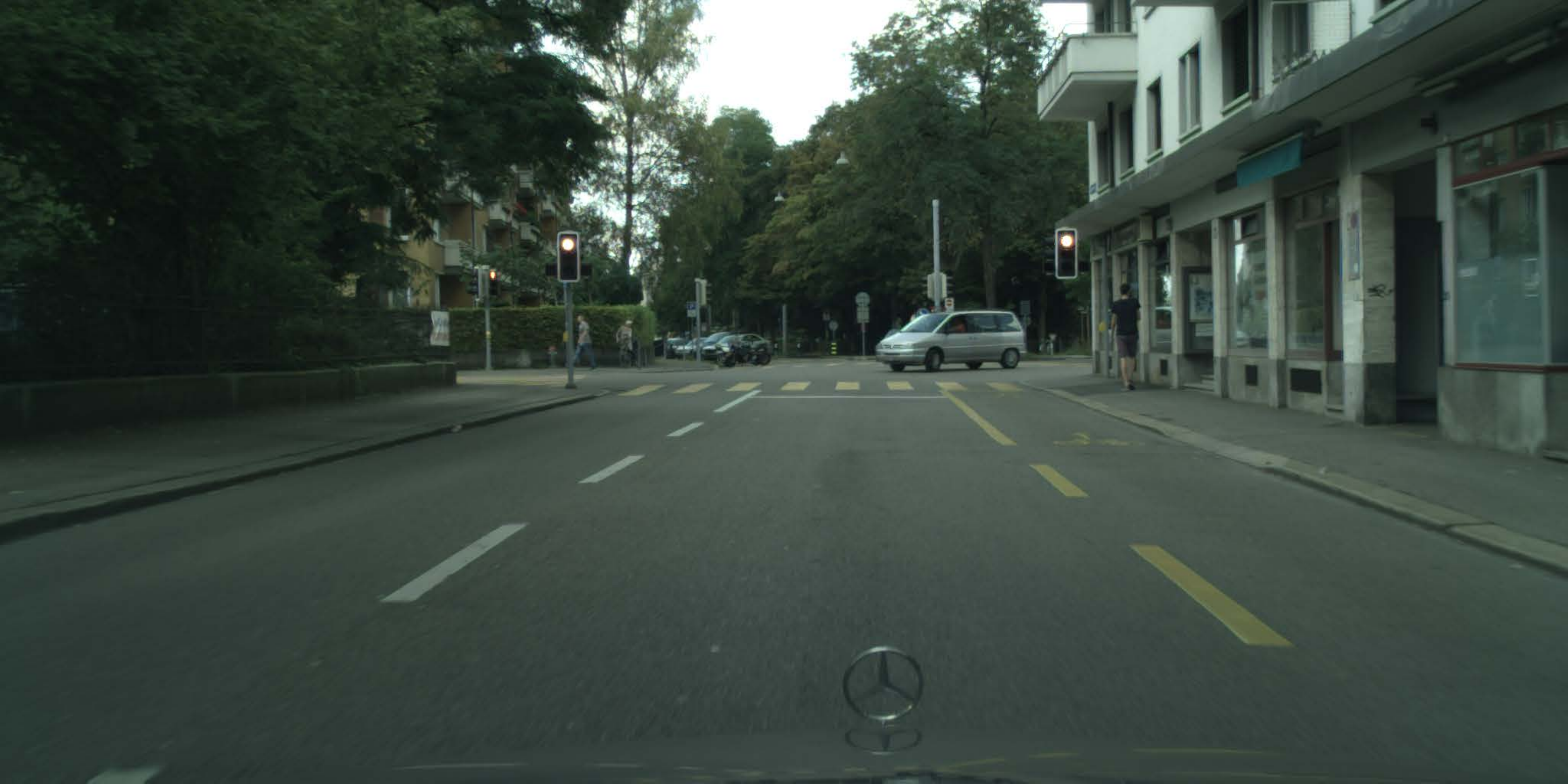}
        \caption{Target image}\label{fig_motivation1}
      \end{subfigure}
      \begin{subfigure}{0.22\textwidth}
        \includegraphics[width=\linewidth]{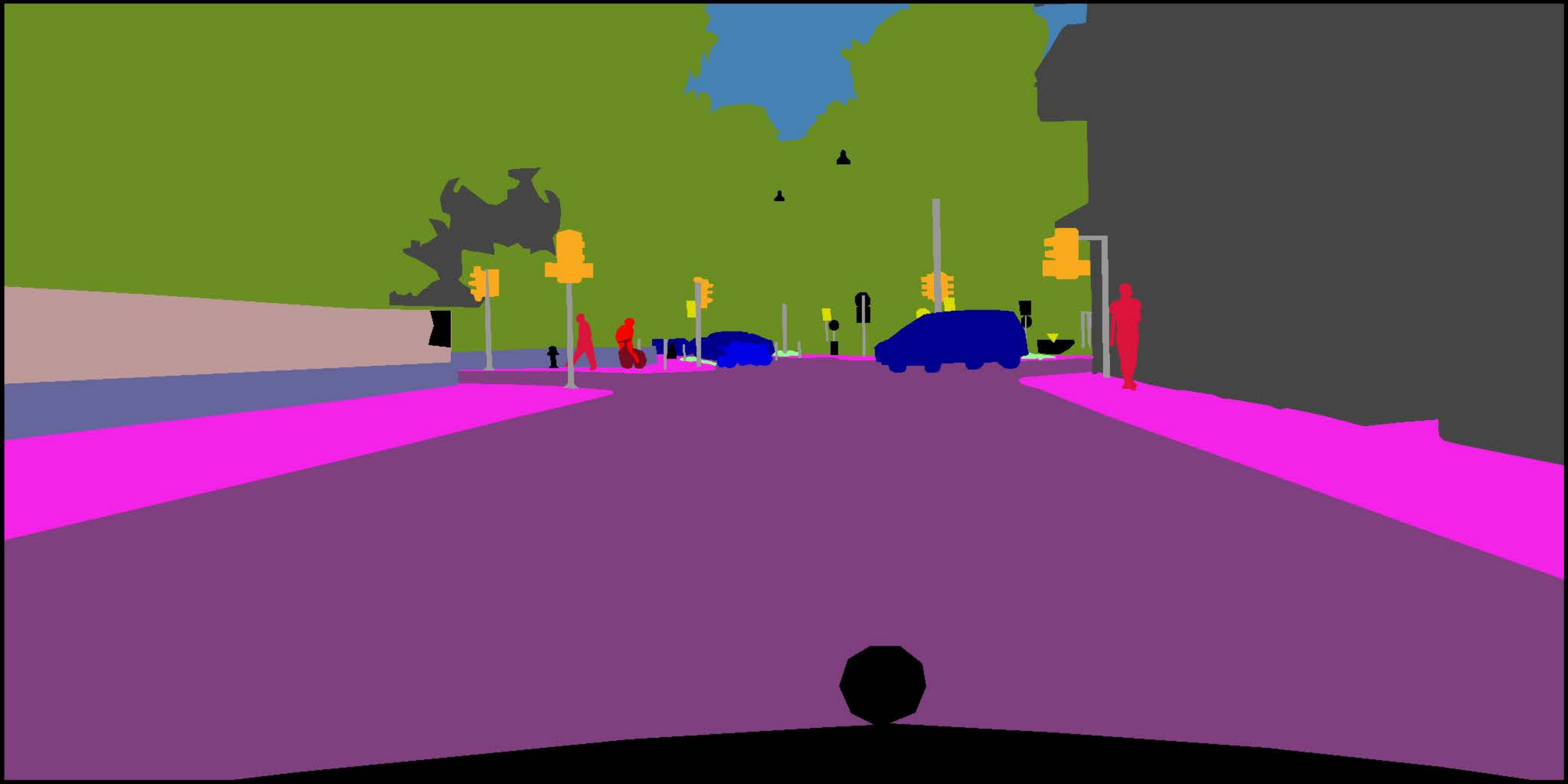}
        \caption{Image-based selection (100\%)}\label{fig_motivation2}
      \end{subfigure}
      \begin{subfigure}{0.22\textwidth}
        \includegraphics[width=\linewidth]{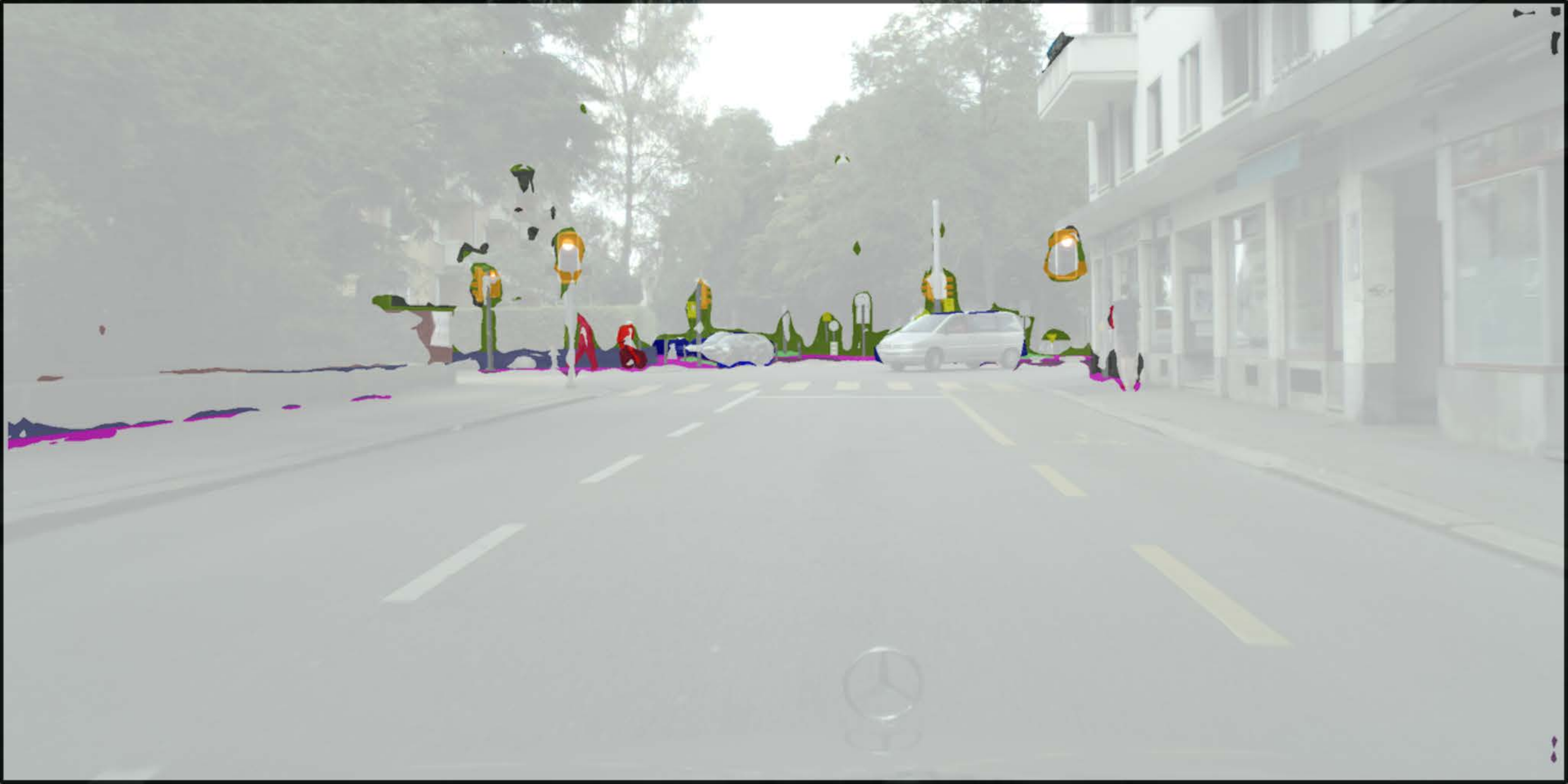}
        \caption{Point-based selection (2.2\%)}\label{fig_motivation3}
      \end{subfigure}
      \begin{subfigure}{0.22\textwidth}
        \includegraphics[width=\linewidth]{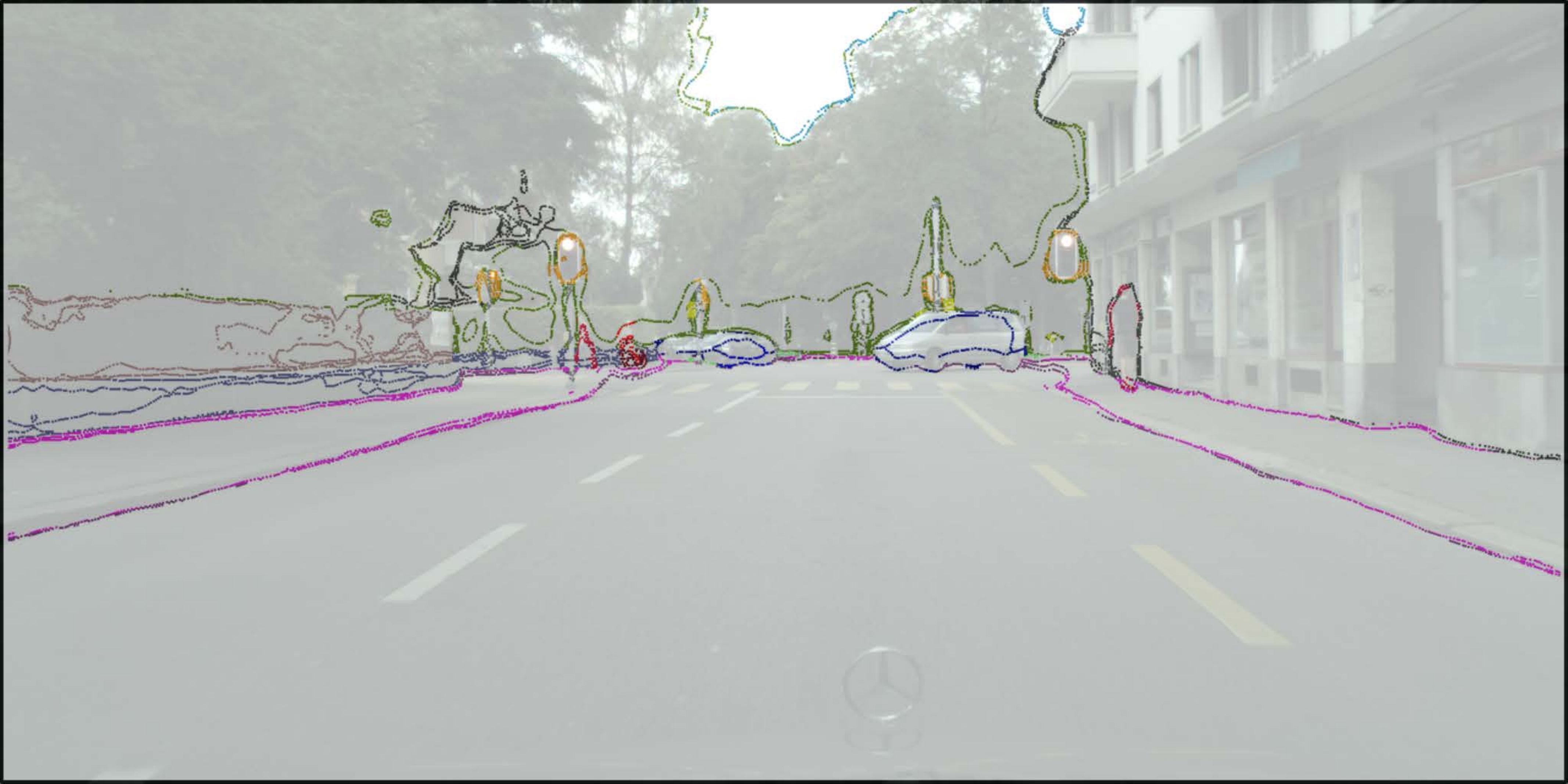}
        \caption{Region-based selection (2.2\%)}\label{fig_motivation4}
      \end{subfigure}\vspace{-2mm}
    \caption{Illustration of different selection strategies. {\bf Image-based selection} (e.g., MADA~\cite{MADA_2021_ICCV}) picks a few target samples and label the entire image, which is probably inefficient. {\bf Point-based selection} (e.g., LabOR~\cite{LabOR_2021_ICCV}) chooses scarce points about which the model is uncertain, while uncertainty estimation at point level is prone to lump pixels that come from particular categories. {\bf Our region-based selection} asks for more annotations of regions with more categories as well as object boundaries in an effective way.} \label{Fig_motivation}
    \vspace{-4mm}
\end{figure}
%%%%%%%%%%%%%%%%%%%%%%%%%%%%%%%%%%%%%%%%%%%%%%%%%%%%%%%%%%%%%%%%%%%%%%%%%%%%%%%%%%%%%%%%%%%%%%%%%%%%%%%%%%%%%%%%%%%%%%%%%%%%%

Semantic segmentation, the task of comprehending an image at the pixel level, is the foundation for numerous applications such as autonomous driving~\cite{TeichmannWZCU18,YangYZLY18}, robot manipulation~\cite{ValadaVDB17,SheikhMLSBS20}, and medical analysis~\cite{ronneberger2015u-net,TaghanakiACCH21}. Learning of segmentation models, however, relies heavily on vast quantities of data with pixel-wise annotations, which is onerous and prohibitively expensive~\cite{cityscapes,Long2014Fully_CVPR}. Further, it remains a major challenge to guarantee a good generalization to diverse testing situations. Various research efforts have been directed to address the above issues, with domain adaptation being promising methods~\cite{tsai2018learning,vu2019advent,GaninUAGLLML16,Tzeng_ADDA,GDCAN,CDAN}. 

Recently, self-training has boosted domain adaptation, which retrains the network with the pseudo labels generated from confident predictions on the target domain~\cite{zou2019confidence,ProDA_2021_CVPR,zheng_2021_IJCV,zou2018unsupervised,MeiZZZ20,Wang_2021_ICCV,Cheng_2021_ICCV}. Nevertheless, this competitive approach faces an inherent challenge: class unbalance is usually extreme. For instance, some classes e.g., ``road" and ``building", appear more frequently than others such as ``rider'' and ``light''. Thereby, pseudo labels are noisy and self-training would put heavy emphasis on classes with high frequency and sacrifice the performance on rare classes or small objects, resulting in undesired biases. Consequently, the performance lags far behind the supervised learning counterparts.

To overcome this obstacle and encourage maximizing segmentation performance on the target domain, we show that a simple active learning strategy works well in adaptive semantic segmentation: annotating a small portion of image regions. Until recently, similar efforts have been made by Ning \etal~\cite{MADA_2021_ICCV} and Shin \etal~\cite{LabOR_2021_ICCV}. The former uses multiple anchors to select representative target images to be labeled (Fig.~\ref{fig_motivation2}), which may be highly inefficient since it can waste the annotation budget on labeling redundant areas within objects. The latter utilizes an inconsistent mask of the bi-classifier predictions to query scarce points in each image for annotation (Fig.~\ref{fig_motivation3}). Although this process reduces human labor costs, under a severe domain shift, uncertainty estimation at the point level may be highly miscalibrated~\cite{SnoekOFLNSDRN19} or lead to sampling redundant points from certain categories. Moreover, both works are straightforward extensions of classification methods and weaken the significance of spatial proximity property in an image. 

Driven by the above analysis, we present a simple, effective, and efficient active learning method, \textit{Region Impurity and Prediction Uncertainty (\method)}, to assist domain adaptive semantic segmentation. A key design element of \method is to select the most diverse and uncertain regions in an image (Fig.~\ref{fig_motivation4}), eventually boosting the segmentation performance. 
To be concrete, we first generate the target pseudo labels from the model predictions and excavate all possible regions with the \KSquareNeighbor algorithm. Second, we take the entropy calculated on the percentage of internal pixels belonging to each distinct class as the region impurity score of each region. Finally, combining region impurity with the mean value of prediction uncertainty, i.e., the entropy of pixel prediction, a novel label acquisition strategy that jointly captures diversity and uncertainty is derived. 

In this work, we introduce two labeling mechanisms for each target image, viz., ``Region-based Annotating (RA)" ($\sim$2.2\% ground truth pixels) and ``Pixel-based Annotating (PA)" ($\sim$40 pixels). \region annotates every pixel in the selected regions---high annotation regime, while \pixel places its focus more on the labeling effort efficiency by selecting the center pixel within the region---low annotation regime.  We further exploit local stability to enforce the prediction consistency between a certain pixel and its neighborhood pixels on the source domain and develop a negative learning loss to enhance the discriminative representation learning on the target domain. We demonstrate that our method can not only help the model to achieve near-supervised performance but also reduce human labeling costs dramatically.  

In a nutshell, our contributions can be summarized as: \vspace{-2mm}
\begin{itemize}
    \item We benchmark the performance of prior methods for active domain adaptation regarding semantic segmentation and uncover that methods using image-based or point-based selection strategies are not effective. \vspace{-2mm}
    \item We propose a region-based acquisition strategy for domain adaptive semantic segmentation, termed \method, that utilizes region impurity and prediction uncertainty to identify image regions that are both diverse in spatial adjacency and uncertain in prediction output. \vspace{-2mm}
    \item We experimentally show that, with standard segmentation models, i.e., DeepLab-v2 and DeepLab-v3+, our method brings significant performance gains across two representative domain adaptation benchmarks, i.e., GTAV $\to$ Cityscapes, SYNTHIA $\to$ Cityscapes.
\end{itemize}

%%%%%%%%% RELATED
\section{Related Work}
\label{sec:related}

\paragraph{Domain adaptation (DA)} enables making predictions on an unlabeled target domain with the knowledge of a well-labeled source domain, which has been widely applied into an array of tasks such as classification~\cite{JADA,AFN,ShuBNE18,Tzeng_ADDA,CDAN,TSA,LiangHF20}, detection~\cite{Vibashan_2021_CVPR,Chen0SDG18} and segmentation~\cite{SFDA_2021_CVPR,0002MPZLYG20}. Initial studies minimize the discrepancy between source and target features to mitigate the domain gap~\cite{DAN-PAMI,Tzeng_DDC,CAN_2019_CVPR}. As to semantic segmentation, most methods employ adversarial learning in three ways: appearance transfer~\cite{li2019bidirectional,Hoffman_cycada2017,yang2020fda}, feature matching~\cite{wang2020differential,zhang2019category,kang2020pixel} and output space alignment~\cite{AdvEnt,luo2019taking,tsai2018learning}. 

Self-training has been gaining momentum as a competitive alternative, which trains the model with pseudo labels on the target domain~\cite{zou2019confidence,Wang_2021_ICCV,ProDA_2021_CVPR,zou2018unsupervised,zheng_2021_IJCV,shin2020two-phase,MeiZZZ20,Cheng_2021_ICCV}. Popular as they are, the pseudo labels are noisy and rely primarily on a good initialization. Some efforts explore additional supervision to engage in this transfer. For example, Paul \etal~\cite{Paul2020WeakSegDA} propose to use weak labels and Vu \etal~\cite{vu2019depth} exploit dense depth information to perform adaptation. Another promising strategy to prevent such noise with minimal annotation workload is active learning, which we adopt in this work.

\paragraph{Active learning (AL)} seeks to minimize labeling effort on an enormous dataset while maximizing performance of the model. Common strategies include uncertainty sampling~\cite{BvSB_2009_CVPR,shen_2018_ICLR,hanneke2014theory} and representative sampling~\cite{BADGE_2020_ICLR,CoreSet_2019_ICLR,sinha_2019_ICCV}. While label acquisition for dense prediction tasks such as segmentation is more expensive and laborious than image classification, there has been considerably less work~\cite{casanove_2020_ICLR,cai_2021_CVPR,kasarla_2019_WACV,siddiqui_2020_CVPR,Wu_2021_ICCV,Shin2021_pixelpick}. A recent example in~\cite{casanove_2020_ICLR} proposes to actively select image regions based on reinforcement learning, which is a more efficient way than labeling entire images.

Up to now, rather little work has been done to consider transferring annotations from a model trained on a given domain (a synthetic dataset) to a different domain (a real-world dataset) due to the dataset shift. However, it occurs frequently in practice but is not adequately addressed. In this work, we take a step forward to deal with this problem.

\paragraph{Active domain adaptation (ADA).} Existing works mainly focus on image classification~\cite{rai2010domain,CULE_2021_ICCV,Fu_2021_CVPR,AADA_WACV,Rangwani_2021_ICCV,EADA}. To name a few, Prabhu \etal~\cite{CULE_2021_ICCV} combine the uncertainty and diversity into an acquisition round and integrate semi-supervised domain adaptation into a unified framework. Lately, Ning \etal~\cite{MADA_2021_ICCV} and Shin \etal~\cite{LabOR_2021_ICCV} are among the first to study the task of ADA applied to semantic segmentation, which greatly enhances the segmentation performance on the target domain. Ning \etal~\cite{MADA_2021_ICCV} put forward a multi-anchor strategy to actively select a subset of images and annotate the entire image, which is probably inefficient. While Shin \etal~\cite{LabOR_2021_ICCV} present a more efficient point-based annotation with an adaptive pixel selector. But, the selected points are individual and discrete, neglecting the contextual structures of an image and pixel spatial contiguity within a region.

Though impressive, these methods neglect the value of spatial adjacency property within a region, and we argue that an effective and efficient region-based selection strategy is essential for lowering human labor costs while preserving model performance. In this work, we explore the spatial coherency of an image and favor the selection of the most diverse and uncertain image regions, promising high information content and low labeling costs.

%%%%%%%%% APPROACH
\section{Approach}
\label{sec:method}

%%%%%%%%%%%%%%%%%%%%%%%%%%%%%%%%%%%%%%%%%%%%%%%%%%%%%%%%%%%%%%%%%%%%%%%%%%%%%%%%%%%%%%%%%%%%%%%%%%%%%%%%%%%%%%%%%%%%%%%%%%%%%
\begin{figure*}[t]
    \centering  
    \includegraphics[width=0.96\textwidth]{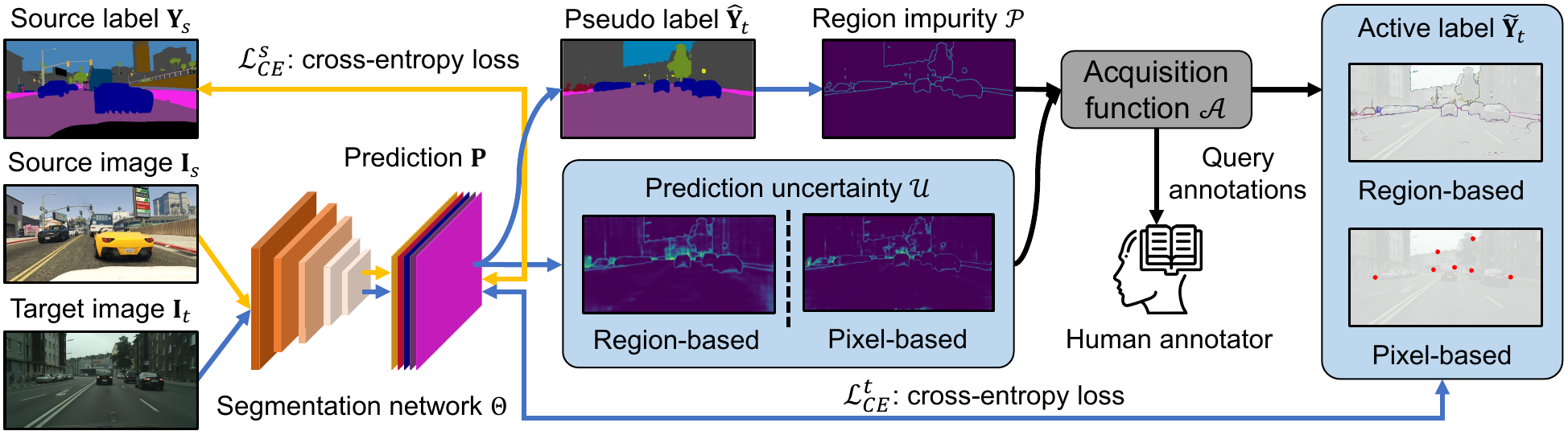} \vspace{-2mm}
    \caption{{\bf The overview of the proposed \method.} At each round of active selection, we first utilize the current model to evaluate the region impurity $\Purity$ and prediction uncertainty $\Uncertainty$ for each unlabeled target. Then we select a batch of regions (\region) or pixels (\pixel) using an acquisition function $\mathcal{A}$, and query annotations from a human annotator. Finally, the network is retrained using all the data labeled so far. 
    } \vspace{-4mm}
    \label{fig:framework}
\end{figure*}
%%%%%%%%%%%%%%%%%%%%%%%%%%%%%%%%%%%%%%%%%%%%%%%%%%%%%%%%%%%%%%%%%%%%%%%%%%%%%%%%%%%%%%%%%%%%%%%%%%%%%%%%%%%%%%%%%%%%%%%%%%%%%

\subsection{Preliminaries and Motivation}
\label{sec:preliminaries}
Formally, in ADA semantic segmentation, we have a set of labeled source data $\mathcal{S}=\{(\Image_s, \Label_s)\}$ and unlabeled target data $\mathcal{T}=\{(\Image_t,\ActiveLabel_t)\}$, where $\Label_s$ is the pixel-wise label belonging to one of the $C$ known classes in label space $\mathcal{Y}$ and $\ActiveLabel_t$ is the target active label that is initialized as $\emptyset$. The goal is to learn a function $h: \Image \to \Label$ (a segmentation network parameterized by $\Theta$) that achieves good segmentation performance on the target domain, with a few annotations. 

Generally, the networks trained on the source domain generalize poorly on the target domain due to domain shift. To effectively transfer knowledge, recent advances resort to self-training techniques~\cite{zou2019confidence,ProDA_2021_CVPR,MeiZZZ20} and optimizes the cross-entropy loss with target pseudo labels $\Pseudo_t$. But, the performance still far under-performs a fully supervised model. We hypothesis that pseudo labels are noisy, thus only the pixels whose prediction confidence is higher than a given threshold are accounted for retraining. In this way, the network training on target images is bootstrapped by the pixels that the model itself is confident in. To address this issue, we propose a simple yet effective active learning approach, Region Impurity and Prediction Uncertainty (\method), to assist domain adaptation by selecting a few informative image regions. The overall framework is illustrated in Fig.~\ref{fig:framework}.

\subsection{Region Generation}
\label{sec:region_generation}
Traditional region-based active learning approaches in semantic segmentation simply divide an image into non-overlapping rectangles~\cite{casanove_2020_ICLR,kasarla_2019_WACV} or employ superpixels algorithms such as SEEDS~\cite{Van_SEEDS_IJCV} to maintain object boundaries~\cite{cai_2021_CVPR,siddiqui_2020_CVPR}. However, we believe that these fixed regions are not flexible or suitable for a region-based selection strategy. The principal reason for this is that the model predictions for adjacent regions should also be considered. 

In this work, we consider \KSquareNeighbor of a pixel as a region, i.e., a regularly-shaped square of size $(2k+1, 2k+1)$ is treated as a region centered on each pixel. Formally, for any pixel $(i, j) \in \mathbb{R}^{H\times W}$ in an image $\Image_t$ with $H$ denoting the heigh and $W$ for width, a region is denoted as
\begin{small}
    \begin{align}
        \label{eq:neighbor}
        \Neighbor_k(i, j) = \{(u, v)~|~|u - i|\le k, |v - j| \le k\} \,.
    \end{align}
\end{small}%
Note that our method is a general one and any other division of region can be employed. In \S~\ref{sec:analysis_results}, we further analyze the effectiveness of the shape and size of a region.

\paragraph{Discussion.} \textit{The concept of region in our work is different from that in recent work}~\cite{HuCW21}. ReginContrast~\cite{HuCW21} is proposed in a supervised manner and ``region'' denotes all pixel features belonging to one class, which is a semantic concept. In contrast, we aim to query informative regions for active domain adaptation and consider a regularly-shaped square centered on each pixel of image as a ``region''.

\subsection{Region Impurity and Prediction Uncertainty}
\label{sec:region_impurity_uncertainty}
We notice that in practice, semantic segmentation often faces class imbalance problem as some events or objects are naturally rare in quantity. Consequently, the performance of the minority classes significantly degrades due to insufficient training samples. One can solve this during the training process through class re-balancing~\cite{WangLKLLTHF20} or augmentation~\cite{He_2021_ICCV}. In contrast, we show that active learning can implicitly solve this by iteratively selecting a batch of samples to annotate at the data collection stage.

Given the pre-defined image regions, we describe our acquisition strategy to implement two different labeling mechanisms for each target image, viz., ``Region-based Annotating (RA)" and ``Pixel-based Annotating (PA)" as follows: at selection iteration $n$, we denote the model as $\Theta^n$, a target image as $\Image_t$ with corresponding active label $\ActiveLabel_t$, and an acquisition function $\ActiveScore(\Image_t; \Theta^n)$ is a function that the active learning system uses to query:
\begin{small}
    \begin{align}
        \label{eq:acquisition_function}
        \mathcal{S}^*=\left\{\begin{aligned}
            & \Neighbor_k(i, j) \; \text{if \region} \\
            & (i, j) \; \text{else \pixel} \\
        \end{aligned}
        \right.
        ,
        (i, j) = \argmax_{(u, v)\notin \ActiveLabel_t} \ActiveScore(\Image_t; \Theta^n)^{(u, v)}.
    \end{align}
\end{small}%
In what follows, we propose an effective acquisition function that jointly captures region impurity and prediction uncertainty, favoring regions that are both diverse in spatial adjacency and uncertain in prediction output.

\paragraph{Region Impurity.} Given a target image $\Image_t$, we first pass it through the network $\Theta$, and then get the softmax output, i.e., prediction $\Probability_{t} \in \mathbb{R}^{H\times W\times C}$. Target pseudo label can be directly derived via the maximum probability output, $\Pseudo_{t}^{(i, j)} = \argmax_{c \in \{1,...,C\}} \Probability_{t}^{(i, j, c)}$. With $\Pseudo_{t}$, we divide a region $\Neighbor_k (i, j)$ of $\Image_t$ into $C$ subsets: 
\begin{small}
    \begin{align}
        \label{eq:division}
        \Neighbor_k^c (i, j) =  \{(u, v) & \in \Neighbor_k(i, j)~|~\Pseudo_{t}^{(u,v)} = c\} \,.
    \end{align}
\end{small}%

At the moment, we can collect statistical information about the categories in a region. If there are many objects in a region, we assume that it helps to train the network after being labeled. Mathematically, we introduce a novel region-based criterion, i.e., region impurity $\Purity$, to assess the significance of regions. Given a region $\Neighbor_k(i, j)$, its region impurity $\Purity^{(i, j)}$ is calculated as 
\begin{small}
    \begin{align}
        \label{eq:purity}
        \Purity^{(i, j)} = - \sumc \frac{|\Neighbor_k^c(i, j)|}{|\Neighbor_k(i, j)|} \log \frac{|\Neighbor_k^c(i, j)|}{|\Neighbor_k(i, j)|} \,,
    \end{align}
\end{small}%%
where $|\cdot |$ denotes the number of pixels in the set.

\paragraph{Prediction Uncertainty.} As the predictions $\mathbf{P}$ carry the semantic relationship information, to measure uncertainty, we employ predictive entropy of each pixel $\Entropy^{(i,j)}$. For $C$-way classification, $\Entropy^{(i, j)} = -\sumc \Probability_{t}^{(i, j, c)} \log \Probability_{t}^{(i, j, c)}$. For one thing, in region-based annotating (\region), we evaluate its prediction uncertainty by the average of all entropies of pixels within the region as follows, 
\begin{small}
    \begin{align}
        \label{eq:region_uncertainty}
        \Uncertainty^{(i, j)} = \frac{1}{|\Neighbor_k(i, j)|} \sum_{(u, v) \in \Neighbor_k(i, j)}\Entropy^{(u,v)} \,.
    \end{align}
\end{small}%
For another, in pixel-based annotating (\pixel), the uncertainty of every pixel is the identity of entropy map, i.e., $\Uncertainty = \Entropy$.

Accordingly, we reckon the final acquisition function as
\begin{small}
    \begin{align}
        \label{eq:active_score}
        \ActiveScore(\Image_t; \Theta^n) = \Purity \odot \Uncertainty\,,
    \end{align}
\end{small}%
where $\odot$ is the element-wise matrix multiplication. 
For \region/\pixel, we query regions/pixels with the highest scores, where the selected regions and the neighbors $\Neighbor_k$ of selected pixels are non-overlapping.

{\bf Discussion.} \textit{Why are region impurity and prediction uncertainty helpful for annotating informative pixels?} First, the region impurity prefers some categories with low occurrence frequency by exploring the spatial adjacency property. As shown in Fig.~\ref{fig:visulization_selected}, this criterion exactly picks out areas of an image with more classes as well as object boundaries. Second, the prediction uncertainty is capable of finding the regions whose model prediction is unsure as well. Finally, these two acquisition criteria gradually select the most diverse and uncertain regions for retraining the model, and in turn, make predictions closer to the ground-truth labels.

%%%%%%%%%%%%%%%%%%%%%%%%%%%%%%%%%%%%%%%%%%%%%%%%%%%%%%%%%%%%%%%%%%%%%%%%%%%%%%%%%%%%%%%%%%%%%%%%%%%%%%%%%%%%%%%%%%%%%%%%%%%%%
\begin{algorithm}[t]
  \DontPrintSemicolon
  {\bf Require}: Labeled source data $(\Image_s, \Label_s)$, unlabeled target data $\Image_t$, total iteration $N$, selection rounds $S$, per-round budget $b$, and hyperparameters: $\tau,\alpha_1,\alpha_2$. \\
  {\bf Define:} Target active label $\ActiveLabel_t=\emptyset$. \\
  Pre-train the model $\Theta^0$ on source data with Eq.~\eqref{eq:loss_seg}. \\
  \For{$n=1 \text{ to } N$}
  {
      Randomly select a batch source data and target and train the network $\Theta^{n}$ via Eq.~\eqref{eq:over_all_loss}. \\ 
      \If{$n \in S$}
      {
              Compute acquisition scores for all regions / pixels in $\Image_t$ according to Eq. \eqref{eq:active_score}. \\
              Iteratively sample regions/pixels using Eq.\eqref{eq:acquisition_function}, and annotate by $\ActiveLabel_{t}[\mathcal{S}^*] = \Label_{t}[\mathcal{S}^*]$, until per-round $b$ is exhausted.
  }
  }
  {\bf Return}: Final model parameters $\Theta^{N}$.
  \caption{Our proposed \method} 
  \label{alg:ours}
\end{algorithm} 
%%%%%%%%%%%%%%%%%%%%%%%%%%%%%%%%%%%%%%%%%%%%%%%%%%%%%%%%%%%%%%%%%%%%%%%%%%%%%%%%%%%%%%%%%%%%%%%%%%%%%%%%%%%%%%%%%%%%%%%%%%%%%

\subsection{Training Objectives}
\label{sec:overall_objective}
With actively selected and annotated regions/pixels in target data, we can train the network to learn information exclusive to the target domain. Therefore, all labeled data from the source and target domain are used to fine-tune the network by optimizing the standard supervised loss:
\begin{small}
    \begin{align}
        \Losssup = \Lossce^s(\Image_s, \Label_s) + \Lossce^t(\Image_t,\ActiveLabel_t) \,,
    \end{align}
\end{small}%
where $\Lossce$ is the categorical cross-entropy (CE) loss:
\begin{small}
    \begin{align}
        \label{eq:loss_seg}
        \Lossce = -\frac{1}{|\Image|}\sums \sumc \Label^{(i, j, c)}\log \Probability^{(i, j, c)}\,,
    \end{align}
\end{small}%
where $\Label^{(i, j, c)}$ denotes the label for pixel $(i,j)$. Meanwhile, we enforce the prediction consistency between a certain pixel and its neighborhood pixels in source images, which will promote the trained model to give more smooth predictions and avoid overfitting to the source data. Formally, this consistency regularization term is formulated as
\begin{small}
    \begin{align}
        \label{eq:loss_soft}
        \Losssoft^s = \frac{1}{|\Image_s|} \sumss \left\lVert \Output^{(i,j)} - \overline{\Output}^{(i, j)}\right\rVert_1 \,,
    \end{align}
\end{small}%
where $\overline{\Output}^{(i, j)}$ is the mean prediction of all pixels in a \textit{1-square-neighbors}, i.e., the region size is 3$\times$3, which can be calculated via $\overline{\Output}^{(i, j)} = \frac{1}{|\Neighbor_1(i, j)|}\sum_{(u,v)\in \Neighbor_1(i, j)} \Output^{(u,v)}$.

Additionally, the lower output probabilities for target images in the early stages of training actually show particular absent classes, called negative pseudo labels~\cite{KimYYK19,Wang_2021_give_your_model}. For instance, it is hard to judge which class a pixel with a predicted value of [0.49, 0.50, 0.01] belongs to, but we can clearly know that it does not belong to the class with a score of 0.01. Thus, we assign negative pseudo labels as below
\begin{small}
    \begin{align}
        \pi(\Probability_{t}^{(i, j, c)}) = \left\{\begin{aligned}
            & 1 \quad \text{if}~\Probability_{t}^{(i, j, c)} < \tau\,, \\
            & 0 \quad \text{otherwise}\,, \\
        \end{aligned}
        \right.
    \end{align}
\end{small}%%
where $\tau$ is the negative threshold, and we use $\tau = 0.05$ in this work. Note that the negative pseudo labels are binary labels. Hence, the negative learning loss is formulated as
\begin{small}
    \begin{align}
        \Lossneg^t  = \frac{-1}{\Lambda(\Image_t)} \sumst \sumc \pi(\Probability_{t}^{(i, j, c)})\log (1-\Probability_{t}^{(i, j, c)}) \label{eq:loss_neg} \,,
    \end{align}
\end{small}% 
where $\Lambda(\Image_t)$ denotes all available negative pseudo labels and is calculated by $\Lambda(\Image_t) = \sumst \sumc \pi(\Probability_{t}^{(i, j, c)})$.

Eventually, equipped with all the above losses, we train the network with the following total objective: 
\begin{small}
    \begin{align}
        \min_{\Theta} \Losssup + \alpha_1\Losssoft^s + \alpha_2\Lossneg^t\,.
        \label{eq:over_all_loss}
    \end{align}
\end{small}%
By default, the weighting coefficients $\alpha_1$ and $\alpha_2$ both are set to 0.1 and 1.0 respectively in all experiments. To be clear, we summarize the overall algorithm in Algorithm~\ref{alg:ours}. 

%%%%%%%%% EXPERIMENT
\section{Experiments}
\label{sec:Experiment}
%%%%%%%%%%%%%%%%%%%%%%%%%%%%%%%%%%%%%%%%%%%%%%%%%%%%%%%%%%%%%%%%%%%%%%%%%%%%%%%%%%%%%%%%%%%%%%%%%%%%%%%%%%%%%%%%%%%%%%%%%%%%%
\begin{table*}[t]
  \centering
  \caption{\textbf{Comparison with previous results on task GTAV $\to$ Cityscapes.} We report the mIoU and best results are shown in \textbf{bold}.} \label{table:gta5_active}
  \vspace{-2mm}
  \resizebox{\textwidth}{!}{
  \begin{threeparttable}
  \begin{tabular}{l c c c c c c c c c c c c c c c c c c c c}
  \toprule[1.2pt]
  Method & \rotatebox{60}{road} & \rotatebox{60}{side.} & \rotatebox{60}{buil.} & \rotatebox{60}{wall} & \rotatebox{60}{fence} & \rotatebox{60}{pole} & \rotatebox{60}{light} & \rotatebox{60}{sign} & \rotatebox{60}{veg.} & \rotatebox{60}{terr.} & \rotatebox{60}{sky} & \rotatebox{60}{pers.} & \rotatebox{60}{rider} & \rotatebox{60}{car} & \rotatebox{60}{truck} & \rotatebox{60}{bus} & \rotatebox{60}{train} & \rotatebox{60}{motor} & \rotatebox{60}{bike} & mIoU\\
  \midrule
  Source Only & 75.8 & 16.8 & 77.2 & 12.5 & 21.0 & 25.5 & 30.1 & 20.1 & 81.3 & 24.6 & 70.3 & 53.8 & 26.4 & 49.9 & 17.2 & 25.9 & 6.5 & 25.3 & 36.0 & 36.6 \\
  CBST~\cite{zou2018unsupervised} & 91.8 & 53.5 & 80.5 & 32.7 & 21.0 & 34.0 & 28.9 & 20.4 & 83.9 & 34.2 & 80.9 & 53.1 & 24.0 & 82.7 & 30.3 & 35.9 & 16.0 & 25.9 & 42.8 & 45.9 \\
  MRKLD~\cite{zou2019confidence} & 91.0 & 55.4 & 80.0 & 33.7 & 21.4 & 37.3 & 32.9 & 24.5 & 85.0 & 34.1 & 80.8 & 57.7 & 24.6 & 84.1 & 27.8 & 30.1 & 26.9 & 26.0 & 42.3 & 47.1 \\
  Seg-Uncertainty~\cite{zheng_2021_IJCV} & 90.4 & 31.2 & 85.1 & 36.9 & 25.6 & 37.5 & 48.8 & 48.5 & 85.3 & 34.8 & 81.1 & 64.4 & 36.8 & 86.3 & 34.9 & 52.2 & 1.7 & 29.0 & 44.6 & 50.3 \\
  TPLD~\cite{shin2020two-phase} & 94.2 & 60.5 & 82.8 & 36.6 & 16.6 & 39.3 & 29.0 & 25.5 & 85.6 & 44.9 & 84.4 & 60.6 & 27.4 & 84.1 & 37.0 & 47.0 & 31.2 & 36.1 & 50.3 & 51.2 \\
  DPL-Dual~\cite{Cheng_2021_ICCV} & 92.8 & 54.4 & 86.2 & 41.6 & 32.7 & 36.4 & 49.0 & 34.0 & 85.8 & 41.3 & 86.0 & 63.2 & 34.2 & 87.2 & 39.3 & 44.5 & 18.7 & 42.6 & 43.1 & 53.3 \\
  ProDA~\cite{ProDA_2021_CVPR} & 87.8 & 56.0 & 79.7 & 46.3 & 44.8 & 45.6 & 53.5 & 53.5 & 88.6 & 45.2 & 82.1 & 70.7 & 39.2 & 88.8 & 45.5 & 59.4 & 1.0 & 48.9 & 56.4 & 57.5 \\
  \midrule
  WeakDA (point)~\cite{Paul2020WeakSegDA} & 94.0 & 62.7 & 86.3 & 36.5 & 32.8 & 38.4 & 44.9 & 51.0 & 86.1 & 43.4 & 87.7 & 66.4 & 36.5 & 87.9 & 44.1 & 58.8 & 23.2 & 35.6 & 55.9 & 56.4  \\
  \midrule
  LabOR (40 pixels)~\cite{LabOR_2021_ICCV} & \bf 96.1 & \bf 71.8 & \bf 88.8 & 47.0 & 46.5 & \bf 42.2 & \bf 53.1 & \bf 60.6 & \bf 89.4 & 55.1 & \bf 91.4 & \bf 70.8 & 44.7 & 90.6 & 56.7 & 47.9 & 39.1 & 47.3 & 62.7 & 63.5 \\
  \bf \cellcolor{Gray}Ours (\pixel, 40 pixels) & \cellcolor{Gray}95.5 & \cellcolor{Gray}69.2 & \cellcolor{Gray}88.2 & \cellcolor{Gray}\bf 48.0 & \cellcolor{Gray}\bf 46.5 & \cellcolor{Gray}36.9 & \cellcolor{Gray}45.2 & \cellcolor{Gray}55.7 & \cellcolor{Gray}88.5 & \cellcolor{Gray}\bf 55.3 & \cellcolor{Gray}90.2 & \cellcolor{Gray}69.2 & \cellcolor{Gray}\bf 46.1 & \cellcolor{Gray}\bf 91.2 & \cellcolor{Gray}\bf 70.7 & \cellcolor{Gray}\bf 73.0 & \cellcolor{Gray}\bf 58.2 & \cellcolor{Gray}\bf 50.1 & \cellcolor{Gray}\bf 65.9 & \cellcolor{Gray}\bf 65.5 \\
  \midrule
  LabOR (2.2\%)~\cite{LabOR_2021_ICCV} & \bf 96.6 & \bf 77.0 & 89.6 & 47.8 & 50.7 & \bf 48.0 & \bf 56.6 & \bf 63.5 & 89.5 & \bf 57.8 & 91.6 & 72.0 & 47.3 & 91.7 & 62.1 & 61.9 & 48.9 & 47.9 & 65.3 & 66.6 \\
  \bf \cellcolor{Gray}Ours (\region, 2.2\%) & \cellcolor{Gray}96.5 & \cellcolor{Gray}74.1 & \cellcolor{Gray}\bf 89.7 & \cellcolor{Gray}\bf 53.1 & \cellcolor{Gray}\bf 51.0 & \cellcolor{Gray}43.8 & \cellcolor{Gray}53.4 & \cellcolor{Gray}62.2 & \cellcolor{Gray}\bf 90.0 & \cellcolor{Gray}57.6 & \cellcolor{Gray}\bf 92.6 & \cellcolor{Gray}\bf 73.0 & \cellcolor{Gray}\bf 53.0 & \cellcolor{Gray}\bf 92.8 & \cellcolor{Gray}\bf 73.8 & \cellcolor{Gray}\bf 78.5 & \cellcolor{Gray}\bf 62.0 & \cellcolor{Gray}\bf 55.6 & \cellcolor{Gray}\bf 70.0 & \cellcolor{Gray}\bf 69.6 \\  
  Fully Supervised (100\%) & 96.8 & 77.5 & 90.0 & 53.5 & 51.5 & 47.6 & 55.6 & 62.9 & 90.2 & 58.2 & 92.3 & 73.7 & 52.3 & 92.4 & 74.3 & 77.1 & 64.5 & 52.4 & 70.1 & 70.2 \\
  \hline
  \hline
  AADA (5\%)$^{\sharp}$~\cite{AADA_WACV} & 92.2 & 59.9 & 87.3 & 36.4 & 45.7 & 46.1 & 50.6 & 59.5 & 88.3 & 44.0 & 90.2 & 69.7 & 38.2 & 90.0 & 55.3 & 45.1 & 32.0 & 32.6 & 62.9 & 59.3 \\
  MADA (5\%)$^{\sharp}$~\cite{MADA_2021_ICCV} &  95.1 &  69.8 &  88.5 &  43.3 &  48.7 &45.7 &  53.3 &59.2 &  89.1 &  46.7 &  91.5 &  73.9 &  50.1 &  91.2 &  60.6 &  56.9 &  48.4 &  51.6 &  68.7 &  64.9 \\
  \bf \cellcolor{Gray}Ours (RA, 5\%)$^{\sharp}$ & \cellcolor{Gray}\bf 97.0 & \cellcolor{Gray}\bf 77.3 & \cellcolor{Gray}\bf 90.4 & \cellcolor{Gray}\bf 54.6 & \cellcolor{Gray}\bf 53.2 & \cellcolor{Gray}\bf 47.7 & \cellcolor{Gray}\bf 55.9 & \cellcolor{Gray}\bf 64.1 & \cellcolor{Gray}\bf 90.2 & \cellcolor{Gray}\bf 59.2 & \cellcolor{Gray}\bf 93.2 & \cellcolor{Gray}\bf 75.0 & \cellcolor{Gray}\bf 54.8 & \cellcolor{Gray}\bf 92.7 & \cellcolor{Gray}\bf 73.0 & \cellcolor{Gray}\bf 79.7 & \cellcolor{Gray}\bf 68.9 & \cellcolor{Gray}\bf 55.5 & \cellcolor{Gray}\bf 70.3 & \cellcolor{Gray}\bf 71.2 \\
  Fully Supervised (100\%)$^{\sharp}$ & 97.4 & 77.9 & 91.1 & 54.9 & 53.7 & 51.9 & 57.9 & 64.7 & 91.1 & 57.8 & 93.2 & 74.7 & 54.8 & 93.6 & 76.4 & 79.3 & 67.8 & 55.6 & 71.3 & 71.9 \\
  \bottomrule[1.2pt]
  \end{tabular}
  \begin{tablenotes}
      \item Methods with $^{\sharp}$ are based on DeepLab-v3+~\cite{chen2018deeplabv3plus} and others are based on DeepLab-v2~\cite{chen2018deeplab} for a fair comparison.
    \end{tablenotes}
  \end{threeparttable}
  }
  \vspace{-3mm}
\end{table*}
%%%%%%%%%%%%%%%%%%%%%%%%%%%%%%%%%%%%%%%%%%%%%%%%%%%%%%%%%%%%%%%%%%%%%%%%%%%%%%%%%%%%%%%%%%%%%%%%%%%%%%%%%%%%%%%%%%%%%%%%%%%%%

%%%%%%%%%%%%%%%%%%%%%%%%%%%%%%%%%%%%%%%%%%%%%%%%%%%%%%%%%%%%%%%%%%%%%%%%%%%%%%%%%%%%%%%%%%%%%%%%%%%%%%%%%%%%%%%%%%%%%%%%%%%%%
\begin{table*}[t]
  \centering
  \caption{\textbf{Comparisons with previous results on task SYNTHIA $\to$ Cityscapes.} We report the mIoUs in terms of 13 classes (excluding the ``wall", ``fence", and ``pole") and 16 classes. Best results are shown in \textbf{bold}.} \label{table:syn_acitve}
  \vspace{-2mm}
  \resizebox{\textwidth}{!}{
  \begin{threeparttable}
  \begin{tabular}{l c c c c c c c c c c c c c c c c c c }
  \toprule[1.2pt]
  Method & \rotatebox{60}{road} & \rotatebox{60}{side.} & \rotatebox{60}{buil.} & \rotatebox{60}{wall*} & \rotatebox{60}{fence*} & \rotatebox{60}{pole*} & \rotatebox{60}{light} & \rotatebox{60}{sign} & \rotatebox{60}{veg.}  & \rotatebox{60}{sky} & \rotatebox{60}{pers.} & \rotatebox{60}{rider} & \rotatebox{60}{car}  & \rotatebox{60}{bus}  & \rotatebox{60}{motor} & \rotatebox{60}{bike} & mIoU & mIoU*  \\
  \midrule
  Source Only & 64.3 & 21.3 & 73.1 & 2.4 & 1.1 & 31.4 & 7.0 & 27.7 & 63.1 & 67.6 & 42.2 & 19.9 & 73.1 & 15.3 & 10.5 & 38.9 & 34.9 & 40.3 \\
  CBST~\cite{zou2018unsupervised} & 68.0 & 29.9 & 76.3 & 10.8 & 1.4 & 33.9 & 22.8 & 29.5 & 77.6 & 78.3 & 60.6 & 28.3 & 81.6 & 23.5 & 18.8 & 39.8 & 42.6 & 48.9  \\
  MRKLD~\cite{zou2019confidence} & 67.7 & 32.2 & 73.9 & 10.7 & 1.6 & 37.4 & 22.2 & 31.2 & 80.8 & 80.5 & 60.8 & 29.1 & 82.8 & 25.0  &19.4 & 45.3 & 43.8 & 50.1 \\ 
  DPL-Dual~\cite{Cheng_2021_ICCV} & 87.5 & 45.7 & 82.8 & 13.3 & 0.6 & 33.2 & 22.0 & 20.1 & 83.1 & 86.0 & 56.6 & 21.9 & 83.1 & 40.3 & 29.8 & 45.7 & 47.0 & 54.2 \\
  TPLD~\cite{shin2020two-phase} & 80.9 & 44.3 & 82.2 & 19.9 & 0.3 & 40.6 & 20.5 & 30.1 & 77.2 & 80.9 & 60.6 & 25.5 & 84.8 & 41.1 & 24.7 & 43.7 & 47.3 & 53.5 \\
  Seg-Uncertainty~\cite{zheng_2021_IJCV} & 87.6 & 41.9 & 83.1 & 14.7 & 1.7 & 36.2 & 31.3 & 19.9 & 81.6 & 80.6 & 63.0 & 21.8 & 86.2 & 40.7 & 23.6 & 53.1 & 47.9 & 54.9  \\
  ProDA~\cite{kim_2021_CVPR} & 87.8 & 45.7 & 84.6 & 37.1 & 0.6 & 44.0 & 54.6 & 37.0 & 88.1 & 84.4 & 74.2 & 24.3 & 88.2 & 51.1 & 40.5 & 45.6 & 55.5 & 62.0 \\ 
  \midrule
  WeakDA (point)~\cite{Paul2020WeakSegDA} & 94.9 & 63.2 & 85.0 & 27.3 & 24.2 & 34.9 & 37.3 & 50.8 & 84.4 & 88.2 & 60.6 & 36.3 & 86.4 & 43.2 & 36.5 & 61.3 & 57.2 & 63.7 \\
  \midrule
  \bf \cellcolor{Gray}Ours (\pixel, 40 pixels) & \cellcolor{Gray}95.8 & \cellcolor{Gray}71.9 & \cellcolor{Gray}87.8 & \cellcolor{Gray}39.9 & \cellcolor{Gray}41.5 & \cellcolor{Gray}38.3 & \cellcolor{Gray}47.1 & \cellcolor{Gray}54.2 & \cellcolor{Gray}89.2 & \cellcolor{Gray}90.8 & \cellcolor{Gray}69.9 & \cellcolor{Gray}48.5 & \cellcolor{Gray}91.4 & \cellcolor{Gray}71.5 & \cellcolor{Gray}52.2 & \cellcolor{Gray}67.2 & \cellcolor{Gray}66.1 & \cellcolor{Gray}72.1 \\
  \bf \cellcolor{Gray}Ours (\region, 2.2\%) & \bf\cellcolor{Gray} 96.8 & \bf\cellcolor{Gray} 76.6 & \bf\cellcolor{Gray} 89.6 & \bf\cellcolor{Gray} 45.0 & \bf\cellcolor{Gray} 47.7 & \bf\cellcolor{Gray} 45.0 & \bf\cellcolor{Gray} 53.0 & \bf\cellcolor{Gray} 62.5 & \bf\cellcolor{Gray} 90.6 & \bf\cellcolor{Gray} 92.7 & \bf\cellcolor{Gray} 73.0 & \bf\cellcolor{Gray} 52.9 & \bf\cellcolor{Gray} 93.1 & \bf\cellcolor{Gray} 80.5 & \bf\cellcolor{Gray} 52.4 & \bf\cellcolor{Gray} 70.1 & \bf\cellcolor{Gray} 70.1 & \bf\cellcolor{Gray} 75.7 \\
  Fully Supervised (100\%) & 96.7 & 77.8 & 90.2 & 40.1 & 49.8 & 52.2 & 58.5 & 67.6 & 91.7 & 93.8 & 74.9 & 52.0 & 92.6 & 70.5 & 50.6 & 70.2 & 70.6 & 75.9 \\
  \hline
  \hline
  AADA (5\%)$^{\sharp}$~\cite{AADA_WACV} &91.3 &57.6 &86.9 &37.6 &48.3 &45.0 &50.4 &58.5 &88.2 &90.3 &69.4 &37.9 &89.9 &44.5 &32.8 &62.5 &61.9 & 66.2 \\
  MADA (5\%)$^{\sharp}$~\cite{MADA_2021_ICCV} &96.5 &74.6 &88.8 &45.9 &43.8 &46.7 &52.4 &60.5 &89.7 &92.2 &74.1 &51.2 &90.9 &60.3 &52.4 &69.4 &68.1  &73.3  \\
  \bf \cellcolor{Gray}Ours (RA, 5\%)$^{\sharp}$ & \bf\cellcolor{Gray} 97.0 & \bf\cellcolor{Gray} 78.9 & \bf\cellcolor{Gray} 89.9 & \bf\cellcolor{Gray} 47.2 & \bf\cellcolor{Gray} 50.7 & \bf\cellcolor{Gray} 48.5 & \bf\cellcolor{Gray} 55.2 & \bf\cellcolor{Gray} 63.9 & \bf\cellcolor{Gray} 91.1 & \bf\cellcolor{Gray} 93.0 & \bf\cellcolor{Gray} 74.4 & \bf\cellcolor{Gray} 54.1 & \bf\cellcolor{Gray} 92.9 & \bf\cellcolor{Gray} 79.9 & \bf\cellcolor{Gray} 55.3 & \bf\cellcolor{Gray} 71.0 & \bf\cellcolor{Gray} 71.4 & \bf\cellcolor{Gray} 76.7 \\
  Fully Supervised (100\%)$^{\sharp}$ & 97.5 & 81.4 & 90.9 & 48.5 & 51.3 & 53.6 & 59.4 & 68.1 & 91.7 & 93.4 & 75.6 & 51.9 & 93.2 & 75.6 & 52.0 & 71.2 & 72.2 & 77.1 \\
  \bottomrule[1.2pt]
  \end{tabular}
  \begin{tablenotes}
      \item Methods with $^{\sharp}$ are based on DeepLab-v3+~\cite{chen2018deeplabv3plus} and others are based on DeepLab-v2~\cite{chen2018deeplab} for a fair comparison.
    \end{tablenotes}
  \end{threeparttable}
  } 
  \vspace{-3mm}
  \end{table*}
%%%%%%%%%%%%%%%%%%%%%%%%%%%%%%%%%%%%%%%%%%%%%%%%%%%%%%%%%%%%%%%%%%%%%%%%%%%%%%%%%%%%%%%%%%%%%%%%%%%%%%%%%%%%%%%%%%%%%%%%%%%%%

\paragraph{Dataset.} For evaluation, we adapt the segmentation from synthetic images, GTAV~\cite{GTA5} and SYNTHIA~\cite{synthia} datasets, to real scenes, the Cityscapes~\cite{cityscapes} dataset. \textbf{GTAV} contains 24,966 1914$\times$1052 images, sharing 19 classes as Cityscapes. \textbf{SYNTHIA} contains 9,400 1280$\times$760 images, sharing 16 classes. \textbf{Cityscapes} includes high-quality urban scene images with a resolution of 2048$\times$1024. It is split into 2,975 training images and 500 images for validation.

\paragraph{Implementation details.} All experiments are conducted on a Tesla V100 GPU with PyTorch~\cite{paszke2019pytorch}. We adopt DeepLab-v2~\cite{chen2018deeplab} and DeepLab-v3+~\cite{chen2018deeplabv3plus} architectures with ResNet-101~\cite{resnet} pre-trained on ImageNet~\cite{imagenet} as backbone. Regarding the training, we use the SGD optimizer with momentum of 0.9 and weight decay of $5\times10^{-4}$, and employ the ``poly'' learning rate schedule with the initial learning rate at $2.5\times10^{-4}$. For all experiments, we train about 40K iterations with batch size of 2, and source data are resized into 1280$\times$720 while target data are resized into 1280$\times$640. Concerning the hyperparameters, we set $k=1$ for \region and $k=32$ for \pixel, $\tau=0.05$, $\alpha_1=0.1$ and $\alpha_2=1.0$. Readers can refer to Appendix~\ref{appendix:parameters} for more details.

\paragraph{Evaluation metric.} As a common practice~\cite{MADA_2021_ICCV,LabOR_2021_ICCV,tsai2018learning,ProDA_2021_CVPR,MeiZZZ20,zheng_2021_IJCV,zou2019confidence}, we report the mean Intersection-over-Union (mIoU)~\cite{everingham2015the} on the Cityscapes validation set. Specifically, we report the mIoU on the shared 19 classes for GTAV $\to$ Cityscapes and report the results on 13 (mIoU*) and 16 (mIoU) common classes for SYNTHIA $\to$ Cityscapes.

\paragraph{Annotation budget.} The selection process lasts for a total of 5 rounds. For region-based annotating (\region), we select 2.2\% and 5.0\% regions in total compared to LabOR~\cite{LabOR_2021_ICCV} and MADA~\cite{MADA_2021_ICCV}, respectively. Regarding pixel-based annotating (\pixel), we select 40 pixels per image like LabOR~\cite{LabOR_2021_ICCV}.

\paragraph{Fully Supervised baseline.} We operate with the annotations of both the source and target data as the upper bound.

\subsection{Comparisons with the state-of-the-arts}
The results on GTAV $\to$ Cityscapes and SYNTHIA $\to$ Cityscapes are shown in Table~\ref{table:gta5_active} and Table~\ref{table:syn_acitve}, respectively. It can be seen that our method dramatically outperforms prior leading self-training methods. Even though we only use 40 pixels per target image, Ours (\pixel) also shows substantial improvements over the prior breaking record method, i.e., ProDA, which implies that active learning is a promising and complementary solution for domain adaptation. 

For the task of GTAV $\to$ Cityscapes, compared to LabOR~\cite{LabOR_2021_ICCV}, the best baseline model, Ours (\pixel) obtains an improvement of 2.0 mIoU, in the meantime, Ours (\region) exceeds by 3.0 mIoU. Similarly, Ours (\region) is able to easily beat MADA~\cite{MADA_2021_ICCV} and AADA~\cite{AADA_WACV} when using the same backbone (DeepLab-v3+) and same annotation budget (5\%). While for the task of SYNTHIA $\to$ Cityscapes, as expected, both Ours (\pixel) and Ours (\region), are superior to their corresponding state-of-the-art methods.

To sufficiently realize the capacity of our method, we also compare it with the Full Supervised model, which is trained on both source and target domain with all images labeled. It is noteworthy that our method even outperforms Full Supervised with respect to some specific categories such as ``rider'', ``bus'', and ``motor'', suggesting that the proposed method can select marvelous regions to surpass the performance of supervised counterpart. 
 
In a nutshell, the results listed in Table~\ref{table:gta5_active} and Table~\ref{table:syn_acitve} show our method performs favorably against existing ADA and competing DA methods regarding semantic segmentation, and performs comparable to Full supervised model, 
confirming the proposed method is effective and efficient.

%%%%%%%%%%%%%%%%%%%%%%%%%%%%%%%%%%%%%%%%%%%%%%%%%%%%%%%%%%%%%%%%%%%%%%%%%%%%%%%%%%%%%%%%%%%%%%%%%%%%%%%%%%%%%%%%%%%%%%%%%%%%%
\begin{figure*}[!htbp]
      \centering  
      \includegraphics[width=0.98\textwidth]{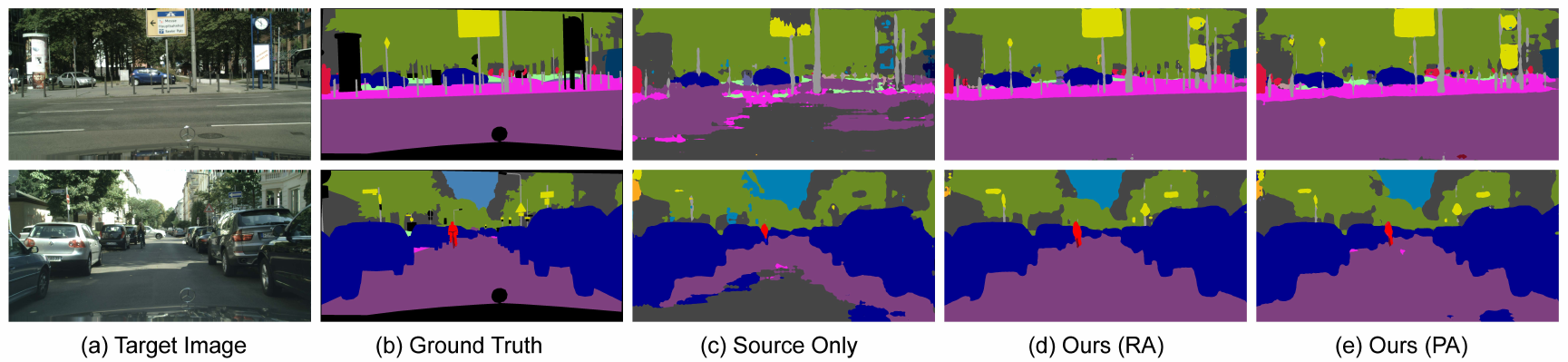}
      \vspace{-2mm}
      \caption{\textbf{Visualization of segmentation results for the task GTAV $\to$ Cityscapes.} From left to right: original target image, ground-truth label, result predicted by Source Only model, result predicted by Ours (\region), and result predicted by Ours (\pixel) are shown one by one.}
      \label{fig:visulization_results}
      \vspace{-2mm}
\end{figure*}%
%%%%%%%%%%%%%%%%%%%%%%%%%%%%%%%%%%%%%%%%%%%%%%%%%%%%%%%%%%%%%%%%%%%%%%%%%%%%%%%%%%%%%%%%%%%%%%%%%%%%%%%%%%%%%%%%%%%%%%%%%%%%%

%%%%%%%%%%%%%%%%%%%%%%%%%%%%%%%%%%%%%%%%%%%%%%%%%%%%%%%%%%%%%%%%%%%%%%%%%%%%%%%%%%%%%%%%%%%%%%%%%%%%%%%%%%%%%%%%%%%%%%%%%%%%%
\begin{figure*}[!htbp]
      \centering  
      \includegraphics[width=0.98\textwidth]{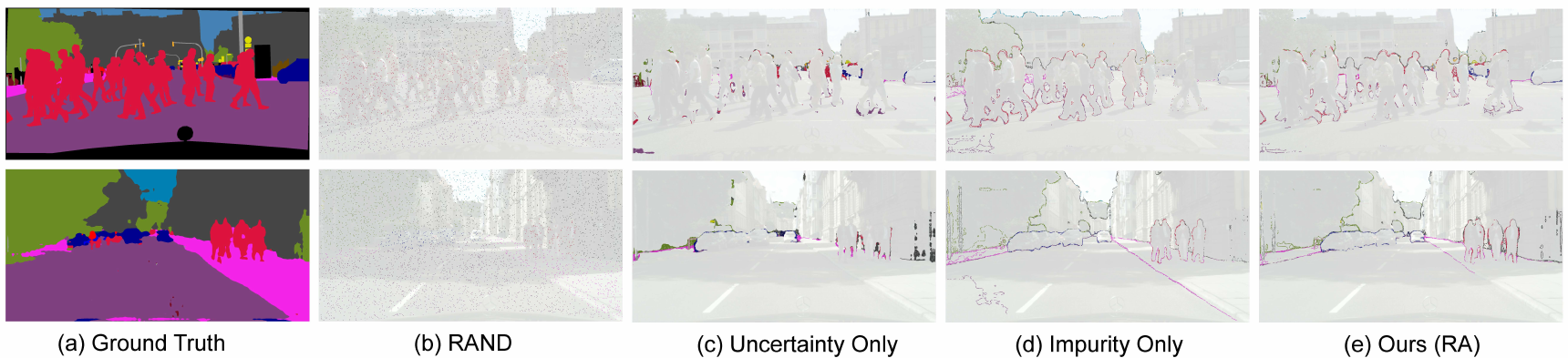}
      \vspace{-2mm}
      \caption{\textbf{Visualization of queried regions to annotate (2.2\%) on GTAV$\to$Cityscapes.} Compared to simple RAND baseline, Uncertainty Only, and Impurity Only, Ours (\region) is able to select the most diverse and uncertain regions of an image. Please zoom in to see the details.}
      \label{fig:visulization_selected}
      \vspace{-2mm}
\end{figure*}
%%%%%%%%%%%%%%%%%%%%%%%%%%%%%%%%%%%%%%%%%%%%%%%%%%%%%%%%%%%%%%%%%%%%%%%%%%%%%%%%%%%%%%%%%%%%%%%%%%%%%%%%%%%%%%%%%%%%%%%%%%%%%

\subsection{Qualitative results}
\label{sec:qualitative_results}
We visualize the segmentation results predicted by \method and compare with those predicted by Source Only model in Fig.~\ref{fig:visulization_results}. The results predicted by \method are smoother and contain less spurious areas than those predicted by the Source Only model, showing that with \method, the performance has been largely improved, especially on hard classes.

Fig.~\ref{fig:visulization_selected} shows the selected regions for annotating from the RAND baseline, Uncertainty Only, Impurity Only, and Ours (\region). We observe that RAND uniformly picks image regions while Uncertainty Only and Impurity Only can cover a larger area of the entire image. However, uncertainty only tends to lump regions that are nearby together and Impurity Only focuses on regions gathered many categories. In contrast, Ours (\region) is also shown to pick diverse regions and not be grounded to a certain region of the image. More importantly, due to the limited budget, we adopt a relative small size to select regions, which might seem like an edge detection visually. Hence, we further operate a simple Canny algorithm~\cite{Canny86a} to select uncertain pixels from pre-detected edge and the results are reported in Appendix~\ref{sec:edge_detection_sampling}.

%%%%%%%%%%%%%%%%%%%%%%%%%%%%%%%%%%%%%%%%%%%%%%%%%%%%%%%%%%%%%%%%%%%%%%%%%%%%%%%%%%%%%%%%%%%%%%%%%%%%%%%%%%%%%%%%%%%%%%%%%%%%%
\begin{table}[t]
  \centering
  \caption{\textbf{Ablation study.} (a): use region impurity only as the selection criterion. (b): use prediction uncertainty only as the selection criterion. (c): combine impurity and uncertainty. (d): train with $\Losssoft^s$ on source samples. (e): train with $\Lossneg^t$ on target samples. (f): our full method \method for region-based annotating.}
  \label{table:ablation}
  \vspace{-2mm}
  \resizebox{\linewidth}{!}{
  \begin{tabular}{c c c c c c c}
  \toprule
   & \multicolumn{2}{c}{Selection} & \multicolumn{2}{c}{Training} & GTAV & SYNTHIA \\
  \midrule
  Method & Impurity & Uncertainty & $\Losssoft^s$ & $\Lossneg^t$ & mIoU & mIoU \\
  \midrule
  \multicolumn{5}{l}{RAND: randomly selecting regions (2.2\%)} & 63.8 & 64.7 \\
  \multicolumn{5}{l}{Fully Supervised: all labeled source and target} & 70.2 & 70.6 \\
  \midrule
  (a) & $\checkmark$ & & & & 68.1 & 69.0 \\
  (b) & & $\checkmark$ & & & 66.2 & 67.9 \\
  (c) & $\checkmark$ & $\checkmark$ & & & 68.5 & 69.2 \\
  \midrule
  (d) & $\checkmark$ & $\checkmark$ & $\checkmark$ &  & 69.0 & 69.7 \\
  (e) & $\checkmark$ & $\checkmark$ & & $\checkmark$ & 69.2 & 69.8 \\
  \cellcolor{Gray}(f) & \cellcolor{Gray}$\checkmark$ & \cellcolor{Gray}$\checkmark$ & \cellcolor{Gray}$\checkmark$ & \cellcolor{Gray}$\checkmark$ & \bf \cellcolor{Gray}69.6 & \bf \cellcolor{Gray}70.1 \\
  \bottomrule
  \end{tabular} 
  }
  \vspace{-2mm}
\end{table}
%%%%%%%%%%%%%%%%%%%%%%%%%%%%%%%%%%%%%%%%%%%%%%%%%%%%%%%%%%%%%%%%%%%%%%%%%%%%%%%%%%%%%%%%%%%%%%%%%%%%%%%%%%%%%%%%%%%%%%%%%%%%%

\subsection{Ablation Study}
\label{sec:ablation}
To further investigate the efficacy of each component of our \method, we perform ablation studies on both GTAV $\to$ Cityscapes and SYNTHIA $\to$ Cityscapes. As shown in Table~\ref{table:ablation}, satisfactory and consistent gains from RAND baseline to our full method prove the effectiveness of each factor. 

\paragraph{Effect of region impurity and prediction uncertainty.} In Table~\ref{table:ablation}, both (a) and (b) achieve a clear improvement compared to RAND. This indicates that the information from either region impurity or prediction uncertainty helps to identify beneficial image regions. Among them, the advantages of the region impurity are more prominent since the spatial adjacency property within an image can alleviate the issue of class imbalance. Furthermore, we can see that (c) is better than both (a) and (b), 
suggesting that integrating the two into a unified selection criterion facilitates the sampling of more diverse and uncertain regions.
riping the benefits of diverse region selection with both impurity and uncertainty

\paragraph{Effect of $\Losssoft^s$ and $\Lossneg^t$.} 
For the training losses, we ablate $\Losssoft^s$ and $\Lossneg^t$ one by one. From the bottom half of Table~\ref{table:ablation}, we can notice that (e) provides 0.7 mIoU gain on GTAV $\to$ Cityscapes and 0.6 on SYNTHIA $\to$ Cityscapes compared to (c). This demonstrates that $\Losssoft^s$ does help to learn local consistent prediction and to avoid overfitting to the source data, which is a complementary factor to the impurity criterion. Similarly, $\Lossneg^t$ on target samples brings a comparable improvement compared to (c). Further, our full \method obtains the best results, which indicate the importance and complementary of the proposed losses. 

%%%%%%%%%%%%%%%%%%%%%%%%%%%%%%%%%%%%%%%%%%%%%%%%%%%%%%%%%%%%%%%%%%%%%%%%%%%%%%%%%%%%%%%%%%%%%%%%%%%%%%%%%%%%%%%%%%%%%%%%%%%%%
\begin{table}[t]
  \centering
  \caption{Experiments on different \textbf{active selection methods}.}\label{table:compare_other_score}
  \vspace{-2mm}
  \resizebox{0.92\linewidth}{!}{
  \begin{tabular}{l c c | c c}
  \toprule
  Method & Budget & mIoU & Budget & mIoU\\
  \midrule
  RAND & 40 pixels & 60.3  & 2.2\% & 63.8 \\
  ENT~\cite{shen_2018_ICLR} & 40 pixels & 55.0  & 2.2\% & 66.2 \\
  SCONF~\cite{culotta_2005_AAAI} & 40 pixels & 59.1 & 2.2\% & 66.5  \\
  \bf \cellcolor{Gray}Ours  & \cellcolor{Gray}\pixel, 40 pixels & \bf \cellcolor{Gray}64.9 & \cellcolor{Gray}\region, 2.2\% & \bf \cellcolor{Gray}68.5 \\  
  \bottomrule
  \end{tabular}
  }
  \vspace{-3mm}
\end{table}
%%%%%%%%%%%%%%%%%%%%%%%%%%%%%%%%%%%%%%%%%%%%%%%%%%%%%%%%%%%%%%%%%%%%%%%%%%%%%%%%%%%%%%%%%%%%%%%%%%%%%%%%%%%%%%%%%%%%%%%%%%%%%

%%%%%%%%%%%%%%%%%%%%%%%%%%%%%%%%%%%%%%%%%%%%%%%%%%%%%%%%%%%%%%%%%%%%%%%%%%%%%%%%%%%%%%%%%%%%%%%%%%%%%%%%%%%%%%%%%%%%%%%%%%%%%
\begin{figure*}[t]
  \centering
	\begin{minipage}[b]{0.665\linewidth} 
      \centering  
      \includegraphics[width=0.98\linewidth]{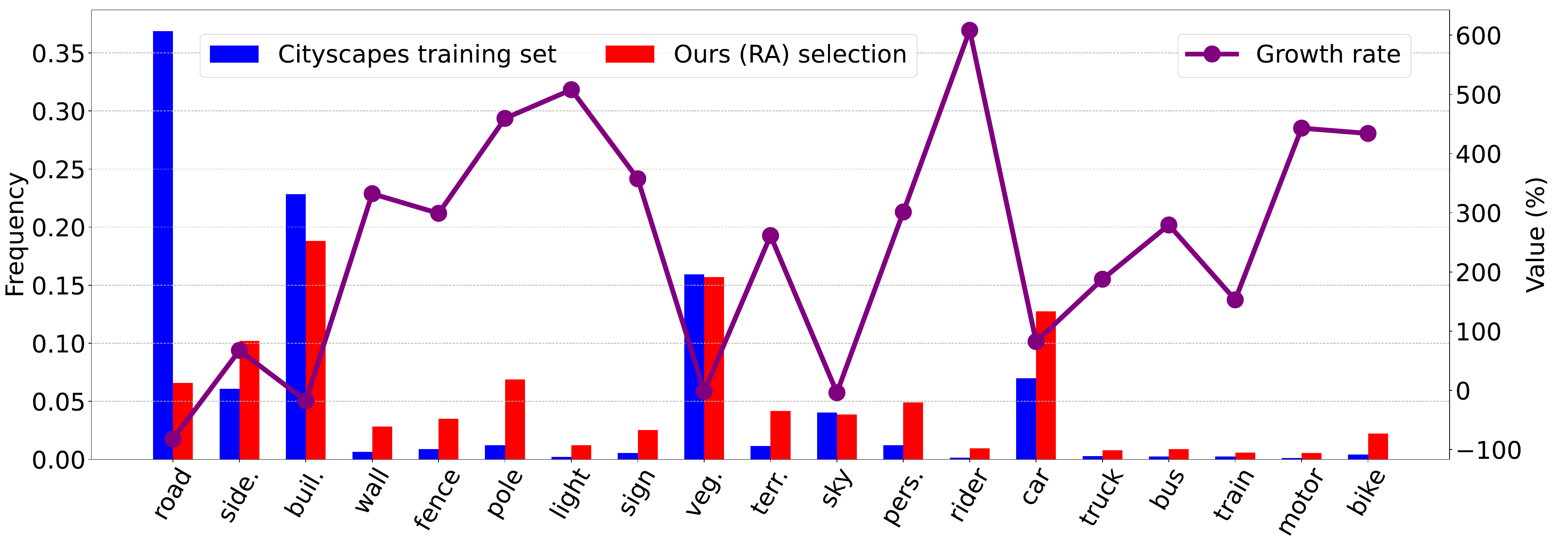} \vspace{-2mm}
      \caption{\textbf{Class frequencies (\%)} in Cityscapes training set for the selected regions via\\ our acquisition strategy. Additionally, we  plot the growth rate between the above two.}
      \label{fig:class_balance}
	\end{minipage}%
	\begin{minipage}[b]{0.37\linewidth} 
      \centering  
      \includegraphics[width=0.98\linewidth]{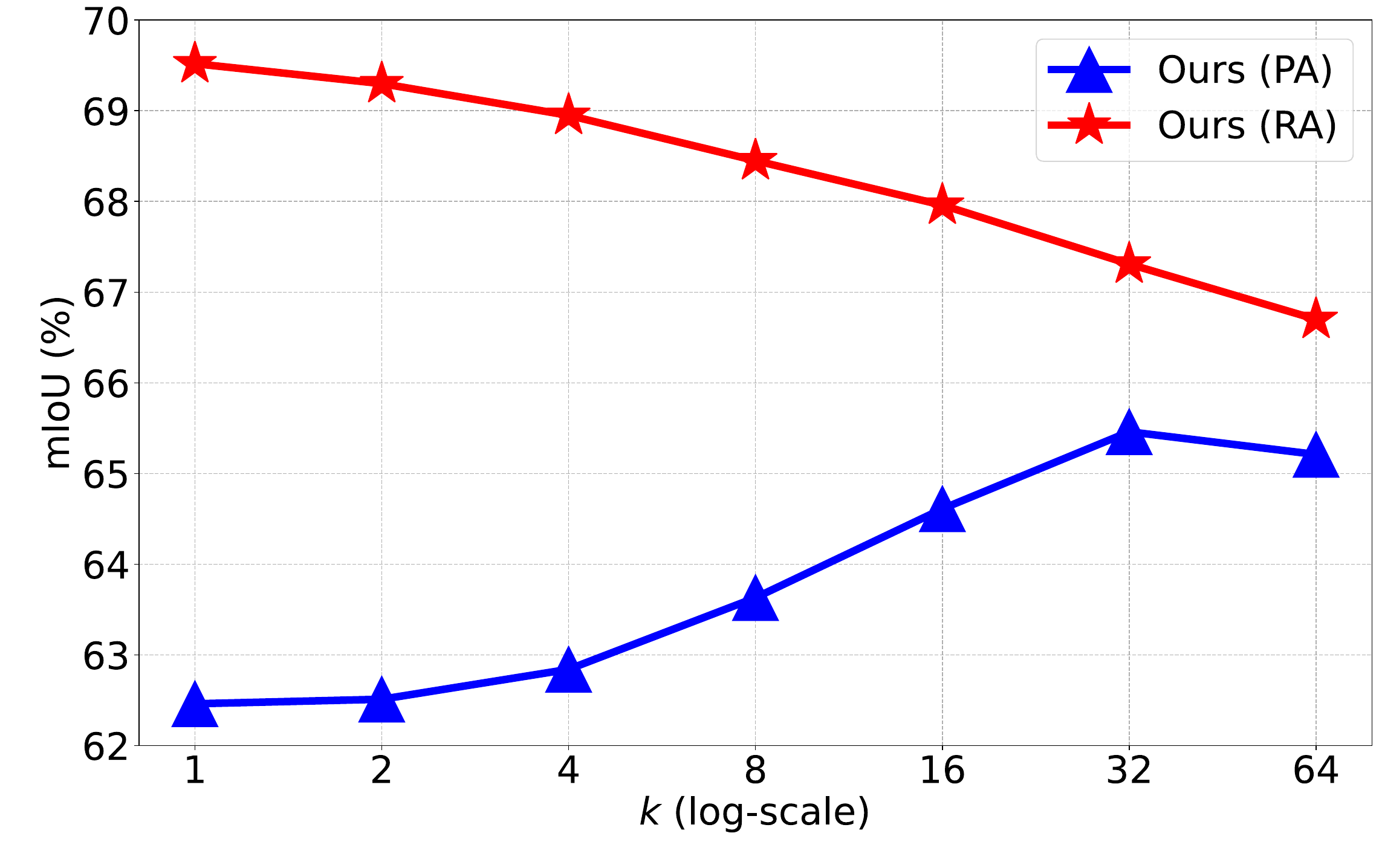} \vspace{-2mm}
      \caption{\textbf{Effect of region size} in the process of region generation on Cityscapes training set.}
      \label{fig:region_size}
	\end{minipage}
\end{figure*} 
%%%%%%%%%%%%%%%%%%%%%%%%%%%%%%%%%%%%%%%%%%%%%%%%%%%%%%%%%%%%%%%%%%%%%%%%%%%%%%%%%%%%%%%%%%%%%%%%%%%%%%%%%%%%%%%%%%%%%%%%%%%%%
\subsection{Further Analysis}
\label{sec:analysis_results}

\paragraph{Comparison of different active selection methods.} To better understand the performance gains from our region-based selection strategy, we compare Ours (\region) and Ours (\pixel), with other common selection methods such as Random selection (RAND), entropy (ENT)~\cite{shen_2018_ICLR} and softmax confidence (SCONF)~\cite{culotta_2005_AAAI}, without $\Losssoft^s$ and $\Lossneg^t$. The results on the task GTAV $\to$ Cityscapes are reported in Table~\ref{table:compare_other_score}. Readers can refer to Appendix~\ref{sec:active_selection_methods} for more details.

Due to a very limited budget of 40 pixels, both ENT and SCONF are severely compromised by the aggregation of selected pixels, and even fall behind the RAND baseline. By comparison, Ours (\pixel) achieves a significant gain over RAND by 4.6 mIoU, proving that our method properly mitigates the effect of redundant pixels to capitalize on a limited budget. Concerning the budget of 2.2\% regions, consistent increases confirm the superiority and advantage.

%%%%%%%%%%%%%%%%%%%%%%%%%%%%%%%%%%%%%%%%%%%%%%%%%%%%%%%%%%%%%%%%%%%%%%%%%%%%%%%%%%%%%%%%%%%%%%%%%%%%%%%%%%%%%%%%%%%%%%%%%%%%%
\begin{table}[t]
  \centering
  \caption{Experiments on \textbf{source-free (SF) scenario}.}\label{table:extension_source_free}
  \vspace{-2mm}
  \resizebox{0.84\linewidth}{!}{
  \begin{tabular}{l c c | c c}
  \toprule[1.2pt]
  \multirow{2}{*}{Method} & \multirow{2}{*}{Budget} & GTAV & \multicolumn{2}{c}{SYNTHIA} \\
  & &  mIoU & mIoU & mIoU* \\
  \midrule
  URMA~\cite{SivaprasadF21} & - & 45.1 & 39.6 & 45.0 \\
  LD~\cite{LD_2021_MM}  & - & 45.5 & 42.6 & 50.1 \\
  SFDA~\cite{GA_SFDA_2021_ICCV} & - & 53.4 & 52.0 & 60.1 \\
  \bf \cellcolor{Gray}Ours (\region) & \cellcolor{Gray}2.2\% & \bf\cellcolor{Gray}67.1 & \bf\cellcolor{Gray}68.7 & \bf \cellcolor{Gray}74.1\\
  \bottomrule[1.2pt]
  \end{tabular}
  }
  \vspace{-2mm}
\end{table}
%%%%%%%%%%%%%%%%%%%%%%%%%%%%%%%%%%%%%%%%%%%%%%%%%%%%%%%%%%%%%%%%%%%%%%%%%%%%%%%%%%%%%%%%%%%%%%%%%%%%%%%%%%%%%%%%%%%%%%%%%%%%%

\paragraph{Extension to source-free scenario.} Considering the data sensitivity and security issues, we further evaluate the generalization of \method by extending it to source-free (SF) scenario~\cite{LiangHF20}. Results in Table~\ref{table:extension_source_free} validate the effectiveness of \method for this challenging DA task. More details about how \method works well are provided in Appendix~\ref{sec:source_free}.

\paragraph{Class frequencies.} Fig.~\ref{fig:class_balance} shows a detailed plot of the class frequencies of the 2.2\% regions selected by Ours (RA) and true class frequencies of the Cityscape training set. As expected, we clearly see that the true distribution is exactly a long-tail distribution while our method is able to pick out more regions that contain rare classes. Particularly, it asks labels for more annotations of ``light'', ``sign'', ``rider'', ``bus'', ``train'' and ``motor''. This, along with the apparent gains in these classes from Table~\ref{table:gta5_active} and Table~\ref{table:syn_acitve}, confirms the diversity and balance of samples selected by our method.

\paragraph{Effect of region size.}
\label{sec:region_size}
To understand how the region size influences our acquisition strategy, we vary \KSquareNeighbor on the task of GTAV $\to$ Cityscapes in Fig.~\ref{fig:region_size}. We can observe two opposite k-dependent laws for Ours (\region) and Ours (\pixel). For \region, the best performance is obtained at a small size $k=1$ since smaller regions have a finer use of the sampling budget while larger regions contain more well-predicted pixels, causing a waste of budget. On the other hand, the best performance of \pixel occurs at a large size $k = 32$. We suppose when we perform PA with a smaller size, we may repeatedly pick pixels in a certain area. To make full use of the limited annotation budget, one should select pixels that cover the whole image. 

%%%%%%%%%%%%%%%%%%%%%%%%%%%%%%%%%%%%%%%%%%%%%%%%%%%%%%%%%%%%%%%%%%%%%%%%%%%%%%%%%%%%%%%%%%%%%%%%%%%%%%%%%%%%%%%%%%%%%%%%%%%%%
\begin{table}[t]
      \centering
      \caption{Effect of \textbf{region shape}.}
      \label{table:analysis_shape}
      \vspace{-2mm}
      \resizebox{0.94\linewidth}{!}{
      \begin{tabular}{c c c}
      \toprule
      Fixed rectangle & Superpixels & \KSquareNeighbor~{\bf (Ours)} \\
      \midrule
      67.7 & 66.8 & \bf 69.6 \\
      \bottomrule
      \end{tabular}
      }\vspace{-2mm}
\end{table}
%%%%%%%%%%%%%%%%%%%%%%%%%%%%%%%%%%%%%%%%%%%%%%%%%%%%%%%%%%%%%%%%%%%%%%%%%%%%%%%%%%%%%%%%%%%%%%%%%%%%%%%%%%%%%%%%%%%%%%%%%%%%%

\paragraph{Effect of region shape}
\label{sec:region_shape}
In \S~\ref{sec:region_generation}, we define \KSquareNeighbor as a region that includes all possible areas within an image, allowing us adaptively select the most diverse and uncertain regions. In Table~\ref{table:analysis_shape}, we compare \KSquareNeighbor with other shapes of regions such as Fixed rectangle and Superpixels (the off-the-shelf SEEDS~\cite{Van_SEEDS_IJCV} algorithm) on the task GTAV $\to$ Cityscapes. As the model is training the importance of each region varies, however, the generated regions of the other two methods are fixed and do not fit well in this case. Thus, we observe the performance degradation using other methods, demonstrating the proposed region generation centered on each pixel is beneficial for acquiring the most concerning  part in an image.

%%%%%%%%% CONCLUSION
\section{Conclusion}
\label{sec:conclusion}
This paper presents Region Impurity and Prediction Uncertainty (\method), an active learning algorithm to deal with performance limitations of domain adaptive semantic segmentation at minimal label cost. We propose a novel region-based acquisition strategy for the selection of limited target regions that are both diverse in spatial contiguity and uncertain under the model. Other than that, we further explore local consistent regularization on the source domain and negative learning on the target domain to advance the acquisition process. Extensive experiments and ablation studies are conducted to verify the effectiveness of the proposed method. Our \method achieves new state-of-the-art results and performs comparably to the supervised counterparts.
We believe that this work will facilitate the development of stronger machine learning system, including active image segmentation~\cite{Shin2021_pixelpick} and universal domain adaptation~\cite{Ma_2021_ICCV}.

\paragraph{Acknowledgements.} This work was supported by the National Natural Science Foundation of China under Grant No. U21A20519 and No. 61902028.
%%%%%%%%%%%%%%%%%%%%%%%%%%%%%%%%%%%%%%%%%%%%%%%%%%%%%%%%%%%%%%%%%%%%%%%%%%%%%%%%%%%%%%%%%%%%%%%%%%%

%%%%%%%%% REFERENCES
% \newpage
{\small\bibliographystyle{ieee_fullname}\bibliography{reference}}

%%%%%%%%%%%%%%%%%%%%%%%%%%%%%%%%%%%%%%%%%%%%%%%%%%%%%%%%%%%%%%%%%%%%%%%%%%%%%%%%%%%%%%%%%%%%%%%%%%
%%%%%%%% APPENDIX
\clearpage
\appendix

\section{Societal impact}
Our method enables a segmentation model trained on a synthetic dataset more generalizable when deploying in a real-world dataset with the limited annotation cost. Thus, our work may have a positive impact on communities to reduce the cost of annotating the out-of-domain data, which is economic and environmental friendliness. We carry out experiments on benchmark datasets and do not notice any societal issues. It does not involve sensitive attributes.

\section{Sensitivity to different hyper-parameters}
\label{appendix:parameters}
We conduct detailed experiments about the sensitivity to different hyper-parameters related to our method. If not specifically mentioned, all the experiments below are carried out based on DeepLab-v2 with the backbone ResNet-101 on GTAV $\to$ Cityscapes using Ours (\region). When we investigate the sensitivity to a specific hyperparameter, other parameters are fixed to the default values, i.e., $\tau$=0.05, $\alpha_1$=0.1 and $\alpha_2$=1.0 for all experiments. 

\subsection{Effect of the negative threshold $\tau$} In Table~\ref{table:hyper_parameters_tau}, we show the results of Ours (\region) with varying $\tau$, i.e., $\tau \in\{0.01,0.02, 0.05,0.08, 0.10, 0.20\}$. Ours (\region) achieves consistent results within a suitable insecure threshold ($\le 0.10$), but will have a performance drop with a large value of $\tau$ like 0.20. 

\subsection{Effect of the consistency regularization loss weight $\alpha_1$} In Table~\ref{table:hyper_parameters_alpha_1}, we show the results of Ours (\region) with varying $\alpha_1$, i.e., $\alpha_1 \in\{0.0,0.05, 0.1,0.2, 0.5, 1.0\}$. Note that when $\alpha_1$=0.0, the model is trained without any consistency constraint on source data. As we can see, the best performance is achieved at $\alpha_1$=0.1. A smaller or larger value of $\alpha_1$ will either induce a weaker or stronger constraint.

\subsection{Effect of the negative learning loss weight $\alpha_2$} In Table~\ref{table:hyper_parameters_alpha_2}, we show the results of Ours (\region) with varying $\alpha_2$, i.e., $\alpha_2 \in\{0.0, 0.1, 0.5, 1.0, 1.5, 2.0\}$. Note that when $\alpha_2$=0.0, the model is trained without negative learning loss on target data. The performance is stable varying $\alpha_2$, which signifies the equal magnitude between negative learning loss and supervised learning loss.

\section{Comparison with a simple edge detector}
\label{sec:edge_detection_sampling}
In \S~\ref{sec:qualitative_results}, since we set $k$=1 that is relative small for \region, the selected regions via Ours (\region) indicate that the edge pixels are mostly favored for labeling. This makes sense due to the criterion favoring regions which have high spatial entropy. Therefore, we further compare our \method with a simple Canny algorithm~\cite{Canny86a} + prediction uncertainty (ENT~\cite{shen_2018_ICLR}) to select uncertain pixels from pre-detected edge and the results are reported in Table~\ref{table:edge_detector}. The performance drops 1.4 mIoU (under 2.2\% budget) and 8.6 mIoU (under 40 pixels budget). 

%%%%%%%%%%%%%%%%%%%%%%%%%%%%%%%%%%%%%%%%%%%%%%%%%%%%%%%%%%%%%%%%%%%%%%%%%%%%%%%%%%%%%%%%%%%%%%%%%%%%%%%%%%%%%%%%%%%%%%%%%%%%%
\begin{table}[t]
  \centering
  \caption{Effect of the negative threshold $\tau$.} \vspace{-2mm}
  \label{table:hyper_parameters_tau}
  \resizebox{0.94\linewidth}{!}{
  \begin{tabular}{c c c c c c c c}
  \toprule
  $\tau$ & 0.01 & 0.02 & \bf \cellcolor{Gray}0.05 & 0.08 & 0.10 & 0.20 \\
  \midrule
  mIoU & 68.81 & 69.54 & \bf \cellcolor{Gray}69.62 & 69.17 & 68.94 & 68.13 \\
  \bottomrule
  \end{tabular}
  }
  \vspace{-2mm}
\end{table}
%%%%%%%%%%%%%%%%%%%%%%%%%%%%%%%%%%%%%%%%%%%%%%%%%%%%%%%%%%%%%%%%%%%%%%%%%%%%%%%%%%%%%%%%%%%%%%%%%%%%%%%%%%%%%%%%%%%%%%%%%%%%%

%%%%%%%%%%%%%%%%%%%%%%%%%%%%%%%%%%%%%%%%%%%%%%%%%%%%%%%%%%%%%%%%%%%%%%%%%%%%%%%%%%%%%%%%%%%%%%%%%%%%%%%%%%%%%%%%%%%%%%%%%%%%%
\begin{table}[t]
  \centering
  \caption{Effect of the consistency regularization loss weight $\alpha_1$.} \vspace{-2mm}
  \label{table:hyper_parameters_alpha_1}
  \resizebox{0.94\linewidth}{!}{
  \begin{tabular}{c c c c c c c c}
  \toprule
  $\alpha_1$ & 0.0 & 0.05 & \bf \cellcolor{Gray}0.1 & 0.2 & 0.5 & 1.0 \\
  \midrule
  mIoU & 69.22 & 69.45 & \bf \cellcolor{Gray}69.62 & 69.61 & 69.58 & 69.39 \\
  \bottomrule
  \end{tabular}
  }
  \vspace{-2mm}
\end{table}
%%%%%%%%%%%%%%%%%%%%%%%%%%%%%%%%%%%%%%%%%%%%%%%%%%%%%%%%%%%%%%%%%%%%%%%%%%%%%%%%%%%%%%%%%%%%%%%%%%%%%%%%%%%%%%%%%%%%%%%%%%%%%

%%%%%%%%%%%%%%%%%%%%%%%%%%%%%%%%%%%%%%%%%%%%%%%%%%%%%%%%%%%%%%%%%%%%%%%%%%%%%%%%%%%%%%%%%%%%%%%%%%%%%%%%%%%%%%%%%%%%%%%%%%%%%
\begin{table}[t]
  \centering
  \caption{Effect of the negative learning loss weight $\alpha_2$.} \vspace{-2mm}
  \label{table:hyper_parameters_alpha_2}
  \resizebox{0.94\linewidth}{!}{
  \begin{tabular}{c c c c c c c c}
  \toprule
  $\alpha_2$ & 0.0 & 0.1 & 0.5 & \bf \cellcolor{Gray}1.0 & 1.5 & 2.0 \\
  \midrule
  mIoU & 69.04 & 69.27 & 69.44 & \bf \cellcolor{Gray}69.62 & 69.36 & 69.16 \\
  \bottomrule
  \end{tabular}
  }
  \vspace{-2mm}
\end{table}
%%%%%%%%%%%%%%%%%%%%%%%%%%%%%%%%%%%%%%%%%%%%%%%%%%%%%%%%%%%%%%%%%%%%%%%%%%%%%%%%%%%%%%%%%%%%%%%%%%%%%%%%%%%%%%%%%%%%%%%%%%%%%

%%%%%%%%%%%%%%%%%%%%%%%%%%%%%%%%%%%%%%%%%%%%%%%%%%%%%%%%%%%%%%%%%%%%%%%%%%%%%%%%%%%%%%%%%%%%%%%%%%%%%%%%%%%%%%%%%%%%%%%%%%%%%
\begin{table}[t]
  \centering
  \caption{Comparison with edge detector on GTAV $\to$ Cityscape.} \vspace{-2mm}
  \label{table:edge_detector}
  \resizebox{0.96\linewidth}{!}{
  \begin{tabular}{c c c| cc}
  \toprule
  Method & Budget & mIoU  & Budget & mIoU \\
  \midrule
  Canny + ENT & 40 pixels & 56.9 & 2.2\% & 68.2 \\
  \bf \cellcolor{Gray}Ours & \cellcolor{Gray}\pixel, 40 pixels & \bf\cellcolor{Gray}65.5 & \cellcolor{Gray}\region, 2.2\% & \bf\cellcolor{Gray}69.6 \\
  \bottomrule
  \end{tabular}
  }
  \vspace{-2mm}
\end{table}
\begin{figure}[t]
  \centering
  \includegraphics[width=0.45\textwidth]{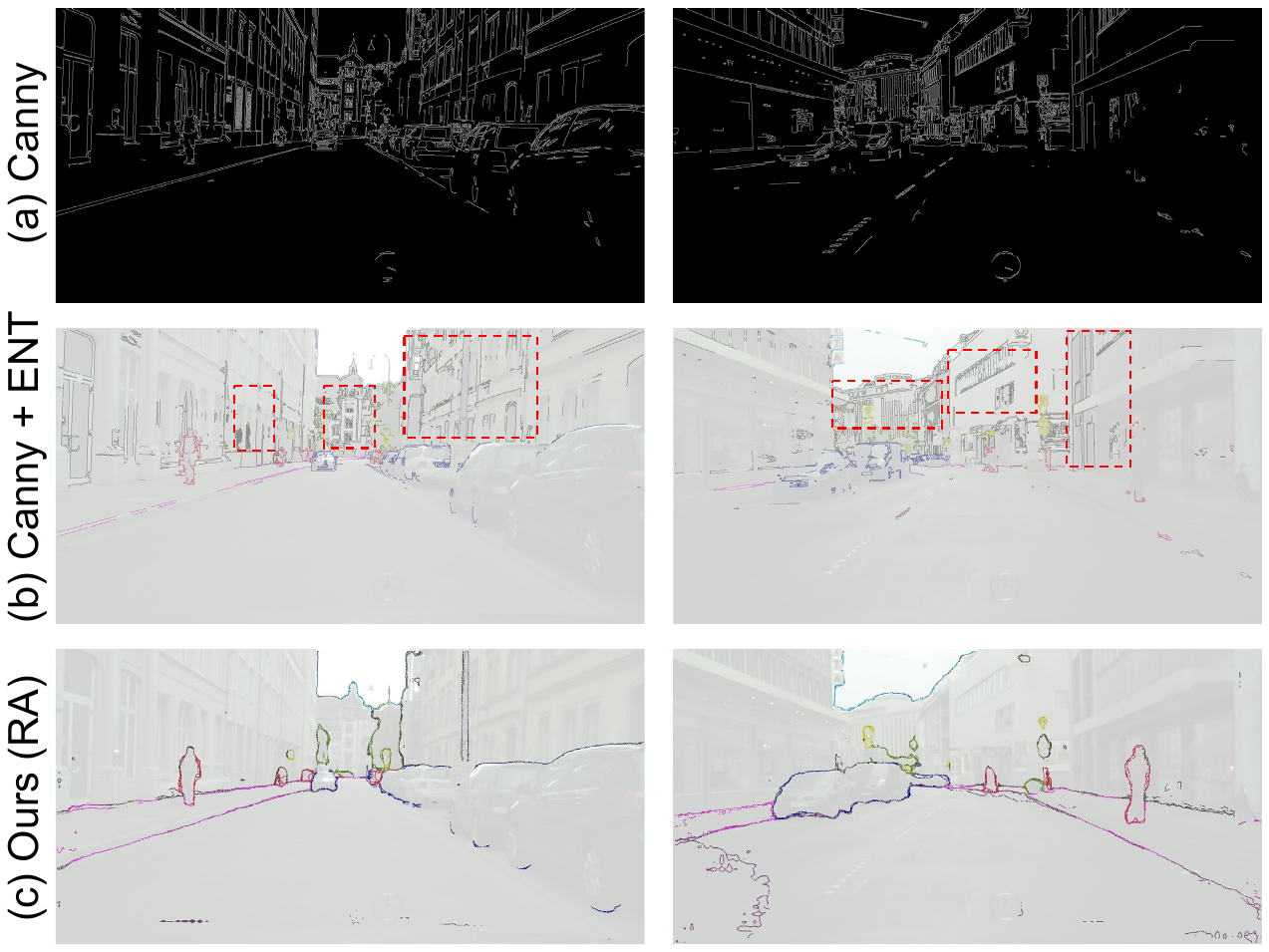} 
  \caption{\textbf{Visualization of queried pixels to annotate (2.2\%) on GTAV $\to$ Cityscapes.} (a) Canny: the edges of target images detected by Canny algorithm; (b) Canny + ENT: a simple Canny edge detector with uncertainty sampling as a baseline. The similarities are that both methods favor the edge pixels to annotate, the differences are that Canny + ENT may pick out pixels inside objects, such as windows in a building while the proposed method is capable of avoiding this issue, demonstrating the benefits of diverse region selection with both impurity and uncertainty.} \label{Fig_compare_canny}
  \vspace{-4mm}
\end{figure}
%%%%%%%%%%%%%%%%%%%%%%%%%%%%%%%%%%%%%%%%%%%%%%%%%%%%%%%%%%%%%%%%%%%%%%%%%%%%%%%%%%%%%%%%%%%%%%%%%%%%%%%%%%%%%%%%%%%%%%%%%%%%%

For a better understanding of the section procedure, we illustrate the selected regions for annotating from the Canny + ENT and Our methods. From Fig.~\ref{Fig_compare_canny}, we can clearly see that the edges inside an object will be selected by Canny + ENT, which is unhelpful. But the region impurity is actually low (low spatial entropy) and these regions would not be selected in our method.

%%%%%%%%%%%%%%%%%%%%%%%%%%%%%%%%%%%%%%%%%%%%%%%%%%%%%%%%%%%%%%%%%%%%%%%%%%%%%%%%%%%%%%%%%%%%%%%%%%%%%%%%%%%%%%%%%%%%%%%%%%%%%
\begin{table*}[!htbp]
  \begin{center}
  \centering
  \caption{Experiments on different active selection methods on GTAV $\to$ Cityscapes. Best results are shown in \textbf{bold}.
  }
  \label{table:compare_other_score_details}
  \vspace{-2mm}
  \resizebox{\textwidth}{!}{
  \begin{tabular}{l | c | c c c c c c c c c c c c c c c c c c r | c}
  \toprule[1.2pt]
  Method & Budget &\rotatebox{60}{road} &\rotatebox{60}{side.} &\rotatebox{60}{buil.} &\rotatebox{60}{wall} &\rotatebox{60}{fence} &\rotatebox{60}{pole} &\rotatebox{60}{light} &\rotatebox{60}{sign} &\rotatebox{60}{veg.} &\rotatebox{60}{terr.} &\rotatebox{60}{sky} &\rotatebox{60}{pers.} &\rotatebox{60}{rider} &\rotatebox{60}{car} &\rotatebox{60}{truck} &\rotatebox{60}{bus} &\rotatebox{60}{train} &\rotatebox{60}{motor} &\rotatebox{60}{bike} & mIoU \\
  \midrule
  RAND & 40 pixels & 94.2 & \bf 69.9 & 85.2 & 44.4 & 40.6 & 33.1 & 41.7 & 49.2 & 85.6 & 51.5 & 88.3 & 62.6 & 37.5 & 87.6 & 61.6 & 62.3 & 49.9 & 41.8 & 59.0 & 60.3  \\
  ENT~\cite{shen_2018_ICLR} & 40 pixels & 93.1 & 55.9 & 85.4 & 35.6 & 30.6 & 30.0 & 28.3 & 39.0 & 86.8 & 45.9 & 88.5 & 65.5 & 32.0 & 88.0 & 55.9 & 54.7 & 27.7 & 39.4 & 62.8 & 55.0  \\
  SCONF~\cite{culotta_2005_AAAI} & 40 pixels & 92.1 & 56.2 & 86.2 & 38.4 & 36.2 & 37.8 & 41.4 & 48.1 & 87.4 & 46.8 & 87.8 & 67.1 & 39.1 & 88.6 & 57.5 & 56.6 & 45.7 & 47.5 & 63.0 & 59.1 \\
  \bf \cellcolor{Gray}Ours (\pixel) & \cellcolor{Gray}40 pixels & \cellcolor{Gray}\bf 95.6 & \cellcolor{Gray}69.6 & \cellcolor{Gray}\bf 88.0 & \cellcolor{Gray}\bf 47.3 & \cellcolor{Gray}\bf 45.1 & \cellcolor{Gray}\bf 37.8 & \cellcolor{Gray}\bf 45.9 & \cellcolor{Gray}\bf 56.5 & \cellcolor{Gray}\bf 88.2 & \cellcolor{Gray}\bf 54.2 & \cellcolor{Gray}\bf 89.0 & \cellcolor{Gray}\bf 69.7 & \cellcolor{Gray}\bf 45.4 & \cellcolor{Gray}\bf 90.9 & \cellcolor{Gray}\bf 67.0 & \cellcolor{Gray}\bf 69.9 & \cellcolor{Gray}\bf 54.1 & \cellcolor{Gray}\bf 52.4 & \cellcolor{Gray}\bf 65.8 & \cellcolor{Gray}\bf 64.9 \\
  \midrule
  RAND & 2.2\% & 95.3 & 72.7 & 86.8 & 45.3 & 43.7 & 38.4 & 45.2 & 53.1 & 87.2 & 54.3 & 90.0 & 65.6 & 42.5 & 59.3 & 67.8 & 67.1 & \bf 59.2 & 45.0 & 63.2 & 63.8  \\
  ENT~\cite{shen_2018_ICLR} & 2.2\% & 94.8 & 71.1 & 87.3 & 52.3 & 46.1 & 38.7 & 47.2 & 56.3 & 87.9 & 55.2 & 89.3 & 69.5 & 47.9 & 90.5 & \bf 74.8 & 71.2 & 58.6 & 52.7 & 66.5 & 66.2 \\
  SCONF~\cite{culotta_2005_AAAI} & 2.2\% & 94.9 & 69.9 & 88.1 & 52.0 & 50.0 & 40.4 & 49.7 & 59.4 & 88.1 & 55.8 & 89.7 & 71.1 & 49.9 & 90.7 & 71.6 & 69.7 & 52.5 & 53.1 & 67.4 & 66.5  \\
  \bf \cellcolor{Gray}Ours (\region) & \cellcolor{Gray}2.2\% & \cellcolor{Gray}\bf 96.9 & \cellcolor{Gray}\bf 76.2 & \cellcolor{Gray}\bf 89.9 & \cellcolor{Gray}\bf 55.5 & \cellcolor{Gray}\bf 52.4 & \cellcolor{Gray}\bf 44.6 & \cellcolor{Gray}\bf 54.5 & \cellcolor{Gray}\bf 63.8 & \cellcolor{Gray}\bf 89.9 & \cellcolor{Gray}\bf 57.0 & \cellcolor{Gray}\bf 92.1 & \cellcolor{Gray}\bf 73.1 & \cellcolor{Gray}\bf 52.7 & \cellcolor{Gray}\bf 92.3 & \cellcolor{Gray}71.9 & \cellcolor{Gray}\bf 72.7 & \cellcolor{Gray}41.0 & \cellcolor{Gray}\bf 55.8 & \cellcolor{Gray}\bf 70.1 & \cellcolor{Gray}\bf 68.5 \\  
  \bottomrule[1.2pt]
  \end{tabular}
  }
  \end{center}
\end{table*}
%%%%%%%%%%%%%%%%%%%%%%%%%%%%%%%%%%%%%%%%%%%%%%%%%%%%%%%%%%%%%%%%%%%%%%%%%%%%%%%%%%%%%%%%%%%%%%%%%%%%%%%%%%%%%%%%%%%%%%%%%%%%%

%%%%%%%%%%%%%%%%%%%%%%%%%%%%%%%%%%%%%%%%%%%%%%%%%%%%%%%%%%%%%%%%%%%%%%%%%%%%%%%%%%%%%%%%%%%%%%%%%%%%%%%%%%%%%%%%%%%%%%%%%%%%%
\begin{figure*}
  \centering  
    \includegraphics[width=0.92\textwidth]{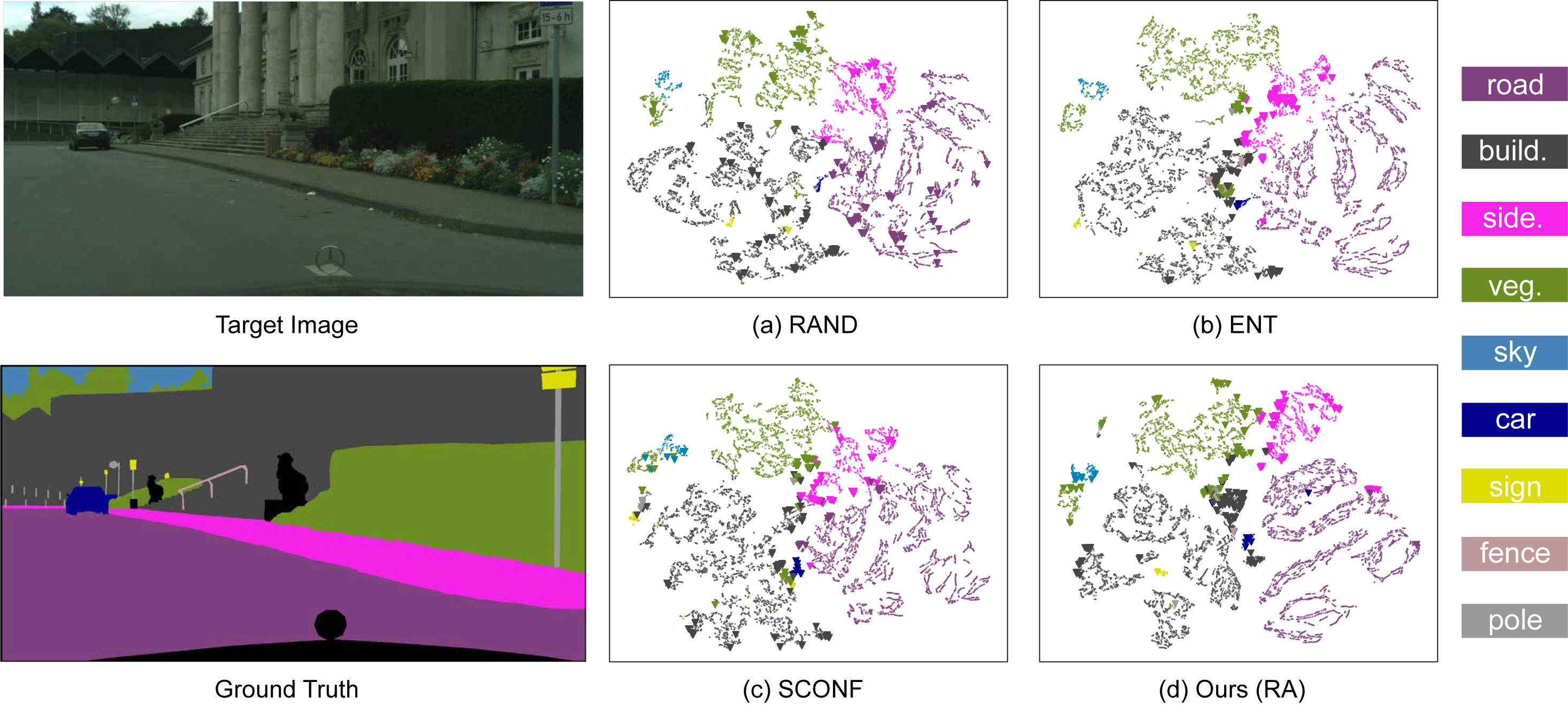}\vspace{-2mm}
    \caption{t-SNE visualization~\cite{tsne} of queried regions from Cityscapes training set on the task GTAV $\to$ Cityscapes. Compared to RAND, ENT, SCONF, Ours (\region) is able to select the most diverse and uncertain regions of an image. 
    Large triangles $\blacktriangledown$ denote the selected pixels and small points $\bullet$ are the remaining pixels in a target image. Please zoom in to see the details.
    } 
    \label{fig:tsne}
\end{figure*}
%%%%%%%%%%%%%%%%%%%%%%%%%%%%%%%%%%%%%%%%%%%%%%%%%%%%%%%%%%%%%%%%%%%%%%%%%%%%%%%%%%%%%%%%%%%%%%%%%%%%%%%%%%%%%%%%%%%%%%%%%%%%%

\section{Comparison of different active selection methods}
\label{sec:active_selection_methods}
\subsection{Details about comparison methods and performance per class}
The performance of active learning depends principally upon depends the selection strategies. In the main paper (Table~\ref{table:compare_other_score}), we compare Ours (RA) and Ours (PA) with other common selection methods such as Random selection (RAND), entropy (ENT)~\cite{shen_2018_ICLR} and softmax confidence (SCONF)~\cite{culotta_2005_AAAI} on GTAV $\to$ Cityscapes. All methods do not use additional loss $\Losssoft^s$ or $\Lossneg^t$. 

\paragraph{Rand selection (RAND):} pixels and regions are randomly sampled with equal probability form each target image.

\paragraph{Entropy (ENT)}~\cite{shen_2018_ICLR}: pixels with the highest prediction entropy, i.e., $-\sumc \Probability_{t}^{(i, j, c)} \log \Probability_{t}^{(i, j, c)}$ are sampled for \pixel. And regions with the highest average prediction entropy of all pixels in a region are sampled for \region.

\paragraph{Softmax confidence (SCONF)}~\cite{culotta_2005_AAAI}: query the most unsure pixels by the softmax confidence score for \pixel
\begin{small}
    \begin{align}
        1 - \max_{c\in\{1,\cdots,C\}} \mathbf{P}_t^{(i, j, c)} \,, \nonumber
    \end{align}
\end{small}%
where larger value indicates less confident. And for \region, select regions with the largest average score of all pixels in a region.

In Table~\ref{table:compare_other_score_details}, we extend the results of Table~\ref{table:compare_other_score} by adding the per class IoU for each method. 
Indeed, our methods select more regions or pixels belonging to the majority classes than baseline methods.
Note that Ours (\region) works specially well for rare object categories, such as ``fence'', ``pole'', ``sign'' or ``rider'', among others, which is a side effect of directly optimizing for per class IoU and mean IoU.

%%%%%%%%%%%%%%%%%%%%%%%%%%%%%%%%%%%%%%%%%%%%%%%%%%%%%%%%%%%%%%%%%%%%%%%%%%%%%%%%%%%%%%%%%%%%%%%%%%%%%%%%%%%%%%%%%%%%%%%%%%%%%
\begin{table*}[!htbp]
  \centering
  \caption{{\bf Quantitative evaluation on GTAV $\to$ Cityscapes.} Results are based on DeepLab-v2 with ResNet-101 architecture. SF indicates whether the method supports source-free adaptation. Best results are shown in {\bf bold}.}\label{table:gtav_source_free}
  \vspace{-2mm}
  \resizebox{\textwidth}{!}{
  \begin{tabular}{l | c | c | c c c c c c c c c c c c c c c c c c r | c}
  \toprule[1.2pt]
  Method & SF & Budget &\rotatebox{60}{road} &\rotatebox{60}{side.} &\rotatebox{60}{buil.} &\rotatebox{60}{wall} &\rotatebox{60}{fence} &\rotatebox{60}{pole} &\rotatebox{60}{light} &\rotatebox{60}{sign} &\rotatebox{60}{veg.} &\rotatebox{60}{terr.} &\rotatebox{60}{sky} &\rotatebox{60}{pers.} &\rotatebox{60}{rider} &\rotatebox{60}{car} &\rotatebox{60}{truck} &\rotatebox{60}{bus} &\rotatebox{60}{train} &\rotatebox{60}{motor} &\rotatebox{60}{bike} & mIoU \\
  \midrule
  URMA~\cite{SivaprasadF21} & \Checkmark & - & 92.3 & 55.2 & 81.6 & 30.8 & 18.8 & 37.1 & 17.7 & 12.1 & 84.2 & 35.9 & 83.8 & 57.7 & 24.1 & 81.7 & 27.5 & 44.3 & 6.9 & 24.1 & 40.4 & 45.1 \\
  LD~\cite{LD_2021_MM} & \Checkmark & - & 91.6 & 53.2 & 80.6 & 36.6 & 14.2 & 26.4 & 31.6 & 22.7 & 83.1 & 42.1 & 79.3 & 57.3 & 26.6 & 82.1 & 41.0 & 50.1 & 0.3 & 25.9 & 19.5 & 45.5 \\
  SFDA (w/ cPAE)~\cite{GA_SFDA_2021_ICCV} & \Checkmark & - & 91.7 & 53.4 & 86.1 & 37.6 & 32.1 & 37.4 & 38.2 & 35.6 & 86.7 & 48.5 & 89.9 & 62.6 & 34.3 & 87.2 & 51.0 & 50.8 & 4.2 & 42.7 & 53.9 & 53.4 \\
  \bf \cellcolor{Gray}Ours (\region) & \cellcolor{Gray}\Checkmark & \cellcolor{Gray}2.2\% & \cellcolor{Gray}95.9 & \cellcolor{Gray}\bf 76.2 & \cellcolor{Gray}88.4 & \cellcolor{Gray}45.4 & \cellcolor{Gray}47.8 & \cellcolor{Gray}42.1 & \cellcolor{Gray}53.0 & \cellcolor{Gray}\bf 62.8 & \cellcolor{Gray}88.6 & \cellcolor{Gray}56.6 & \cellcolor{Gray}91.4 & \cellcolor{Gray}72.1 & \cellcolor{Gray}52.2 & \cellcolor{Gray}91.2 & \cellcolor{Gray}59.5 & \cellcolor{Gray}74.2 & \cellcolor{Gray}55.0 & \cellcolor{Gray}54.4 & \cellcolor{Gray}68.3 & \cellcolor{Gray}67.1 \\
  \midrule
  \bf \cellcolor{Gray}Ours (\region) & \cellcolor{Gray}\XSolidBrush & \cellcolor{Gray}2.2\% & \cellcolor{Gray}\bf 96.5 & \cellcolor{Gray}74.1 & \cellcolor{Gray}\bf 89.7 & \cellcolor{Gray}\bf 53.1 & \cellcolor{Gray}\bf 51.0 & \cellcolor{Gray}\bf 43.8 & \cellcolor{Gray}\bf 53.4 & \cellcolor{Gray}62.2 & \cellcolor{Gray}\bf 90.0 & \cellcolor{Gray}\bf 57.6 & \cellcolor{Gray}\bf 92.6 & \cellcolor{Gray}\bf 73.0 & \cellcolor{Gray}\bf 53.0 & \cellcolor{Gray}\bf 92.8 & \cellcolor{Gray}\bf 73.8 & \cellcolor{Gray}\bf 78.5 & \cellcolor{Gray}\bf 62.0 & \cellcolor{Gray}\bf 55.6 & \cellcolor{Gray}\bf 70.0 & \cellcolor{Gray}\bf 69.6 \\
  \bottomrule[1.2pt]
  \end{tabular}
  }
  \end{table*}
\begin{table*}[!htbp]
  \centering
  \caption{{\bf Quantitative evaluation on SYNTHIA $\to$ Cityscapes.} Results are based on DeepLab-v2 with ResNet-101 architecture. We report the mIoUs in terms of 13 classes (excluding the ``wall", ``fence", and ``pole") and 16 classes. Best results are shown in {\bf bold}.}
  \label{table:syn_source_free}
  \vspace{-2mm}
  \resizebox{\textwidth}{!}{
  \begin{tabular}{l | c | c | c c c c c c c c c c c c c c c c | c c }
  \toprule[1.2pt]
  Method & SF & Budget & \rotatebox{60}{road} &\rotatebox{60}{side.} &\rotatebox{60}{buil.} &\rotatebox{60}{wall*} &\rotatebox{60}{fence*} &\rotatebox{60}{pole*} &\rotatebox{60}{light} &\rotatebox{60}{sign} &\rotatebox{60}{veg.}  &\rotatebox{60}{sky} &\rotatebox{60}{pers.} &\rotatebox{60}{rider} &\rotatebox{60}{car}  &\rotatebox{60}{bus}  &\rotatebox{60}{motor} &\rotatebox{60}{bike} & mIoU & mIoU*  \\
  \midrule
  URMA~\cite{SivaprasadF21} & \Checkmark & - & 59.3 & 24.6 & 77.0 & 14.0 & 1.8 & 31.5 & 18.3 & 32.0 & 83.1 & 80.4 & 46.3 & 17.8 & 76.7 & 17.0 & 18.5 & 34.6 & 39.6 & 45.0 \\
  LD~\cite{LD_2021_MM} & \Checkmark & - & 77.1 & 33.4 & 79.4 & 5.8 & 0.5 & 23.7 & 5.2 & 13.0 & 81.8 & 78.3 & 56.1 & 21.6 & 80.3 & 49.6 & 28.0 & 48.1 & 42.6 & 50.1 \\
  SFDA (w/ cPAE)~\cite{GA_SFDA_2021_ICCV} & \Checkmark & - & 90.5 & 50.0 & 81.6 & 13.3 & 2.8 & 34.7 & 25.7 & 33.1 & 83.8 & 89.2 & 66.0 & 34.9 & 85.3 & 53.4 & 46.1 & 46.6 & 52.0 & 60.1 \\
  \bf \cellcolor{Gray}Ours (\region) & \cellcolor{Gray}\Checkmark & \cellcolor{Gray}2.2\% & \cellcolor{Gray}96.6 & \cellcolor{Gray}75.9 & \cellcolor{Gray}89.0 & \cellcolor{Gray}\bf 49.2 & \cellcolor{Gray}46.6 & \cellcolor{Gray}40.2 & \cellcolor{Gray}48.4 & \cellcolor{Gray}60.7 & \cellcolor{Gray}89.8 & \cellcolor{Gray}91.7 & \cellcolor{Gray}70.0 & \cellcolor{Gray}48.8 & \cellcolor{Gray}92.3 & \cellcolor{Gray}\bf 81.1 & \cellcolor{Gray}51.0 & \cellcolor{Gray}67.8 & \cellcolor{Gray}68.7 & \cellcolor{Gray}74.1 \\
  \midrule
  \bf \cellcolor{Gray}Ours (\region) & \cellcolor{Gray}\XSolidBrush & \cellcolor{Gray}2.2\% & \cellcolor{Gray}\bf 96.8 & \cellcolor{Gray}\bf 76.6 & \cellcolor{Gray}\bf 89.6 & \cellcolor{Gray}45.0 & \cellcolor{Gray}\bf 47.7 & \cellcolor{Gray}\bf 45.0 & \cellcolor{Gray}\bf 53.0 & \cellcolor{Gray}\bf 62.5 & \cellcolor{Gray}\bf 90.6 & \cellcolor{Gray}\bf 92.7 & \cellcolor{Gray}\bf 73.0 & \cellcolor{Gray}\bf 52.9 & \cellcolor{Gray}\bf 93.1 & \cellcolor{Gray}80.5 & \cellcolor{Gray}\bf 52.4 & \cellcolor{Gray}\bf 70.1 & \cellcolor{Gray}\bf 70.1 & \cellcolor{Gray}\bf 75.7 \\
  \bottomrule[1.2pt]
  \end{tabular}
  }
  \end{table*}
%%%%%%%%%%%%%%%%%%%%%%%%%%%%%%%%%%%%%%%%%%%%%%%%%%%%%%%%%%%%%%%%%%%%%%%%%%%%%%%%%%%%%%%%%%%%%%%%%%%%%%%%%%%%%%%%%%%%%%%%%%%%%

\subsection{t-SNE visualization} In Fig.~\ref{fig:tsne}, we illustrate the sampling behavior of Ours (\region) with different selection strategies via t-SNE visualization~\cite{tsne}. We visualize the feature representations of pixels sampled via RAND, ENT, SCONF and Ours (RA) (large, triangles, $\blacktriangledown$) along with the remaining pixels (small, points, $\bullet$) in a target image. We clearly observe that the RAND baseline performs average sampling regions in Fig.~\ref{fig:tsne}(a), which can waste the annotation budget on labeling redundant areas within objects, such as ``road'' and ``building''. Fig.~\ref{fig:tsne}(b) and Fig.~\ref{fig:tsne}(c) show that ENT and SCONF chose most uncertain regions, but seldom chose the infrequent object categories such as ``sign'', ``fence'' and ``pole'' which can be selected via Ours (\region) in Fig.~\ref{fig:tsne}(d). Across all strategies, we find that our method samples regions that are diverse (often present in a region with much more object categories) and uncertain (often present in a cluster of unpredictable object boundaries). 

In short, Ours (\region) is the best option among various possible selection strategies regarding both the performance gain (Table~\ref{table:compare_other_score_details}) and visual presentation (Fig.~\ref{fig:tsne}).

\section{Extension of \method to source-free scenario}
\label{sec:source_free}
Active domain adaptation, which achieves enormous performance gains at the expense of annotating a few target samples, has attracted a surge of interest due to its utility. 
Considering the data sensitivity and security issues, we further evaluate the generalization of our \method to a challenging scenario called source-free domain adaptation (SFDA), where only a source domain pre-trained model and unlabeled target data are accessible to conduct adaptation~\cite{GA_SFDA_2021_ICCV}. In SFDA extension, we start from a source domain pre-trained model, then we optimize the model with $\Lossce^t$ of active samples and $\Lossneg^t$ of target data, without utilizing the source domain. We adopt the DeepLab-v2~\cite{chen2018deeplab} with the backbone ResNet-101 and carry out experiment on both two popular domain adaptation benchmarks, i.e., GTAV $\to$ Cityscapes and SYNTHIA $\to$ Cityscapes, with the annotation budget of 2.2\% regions per image. With respect to training procedure, we keep in line with details of the main paper, such as learning rate schedule, batch size, maximum iteration, and input size etc.

As shown in Table~\ref{table:gtav_source_free} and Table~\ref{table:syn_source_free}, with little workload to manually annotate active regions in a target image, Ours (\region) achieves significant improvements compared to existing SFDA approaches~\cite{GA_SFDA_2021_ICCV,LD_2021_MM,SivaprasadF21}, in detail, 13.7$\sim$22.0 mIoU on GTAV $\to$ Cityscapes and 16.7$\sim$29.1 mIoU on SYNTHIA $\to$ Cityscapes. These results suggest that our method better facilitates the performance on SFDA. In addition, we can observe a slight performance degradation without source data participating during the training process.
In a nutshell, \method can be well generalized to SFDA and shows great potential for further exploration of performance increases.

\section{Additional qualitative results}
\label{sec:additional_qualitative}
\begin{figure*}
  \centering  
    \includegraphics[width=0.98\textwidth]{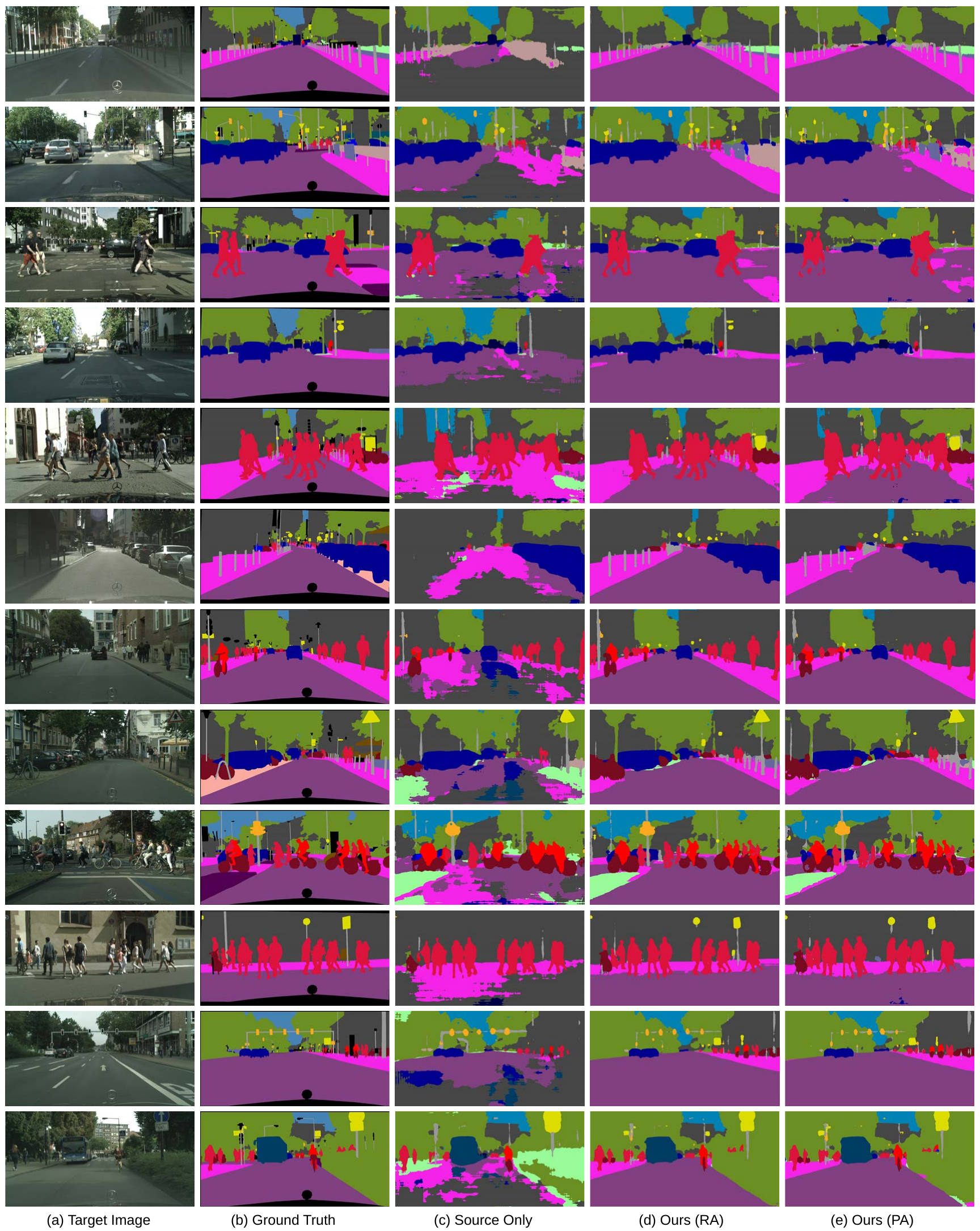} \vspace{-1.5mm}
    \caption{
      {\bf Visualization of segmentation results on GTAV $\to$ Cityscapes.} From left to right: original target image, ground-truth label, result predicted by Source Only model, result predicted by Ours (\region), and result predicted by Ours (\pixel) are shown one by one.
    } \vspace{-4mm}
    \label{fig:results_gta}
\end{figure*}

\begin{figure*}
  \centering  
    \includegraphics[width=0.98\textwidth]{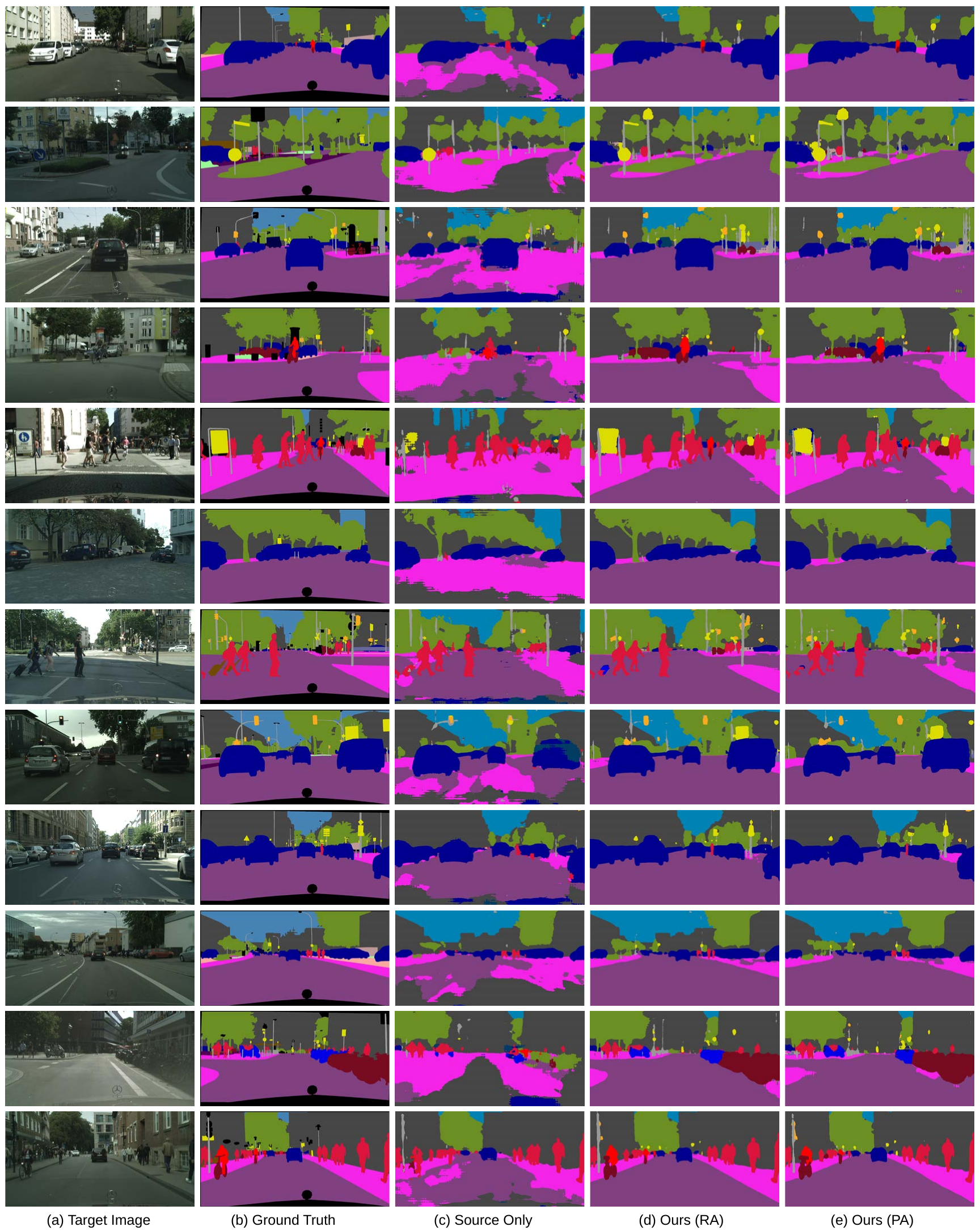} \vspace{-1.5mm}
    \caption{
      {\bf Visualization of segmentation results on SYNTHIA $\to$ Cityscapes.} From left to right: original target image, ground-truth label, result predicted by Source Only model, result predicted by Ours (\region), and result predicted by Ours (\pixel) are shown one by one.
    } \vspace{-4mm}
    \label{fig:results_syn}
\end{figure*}

\begin{figure*}
  \centering  
    \includegraphics[width=0.98\textwidth]{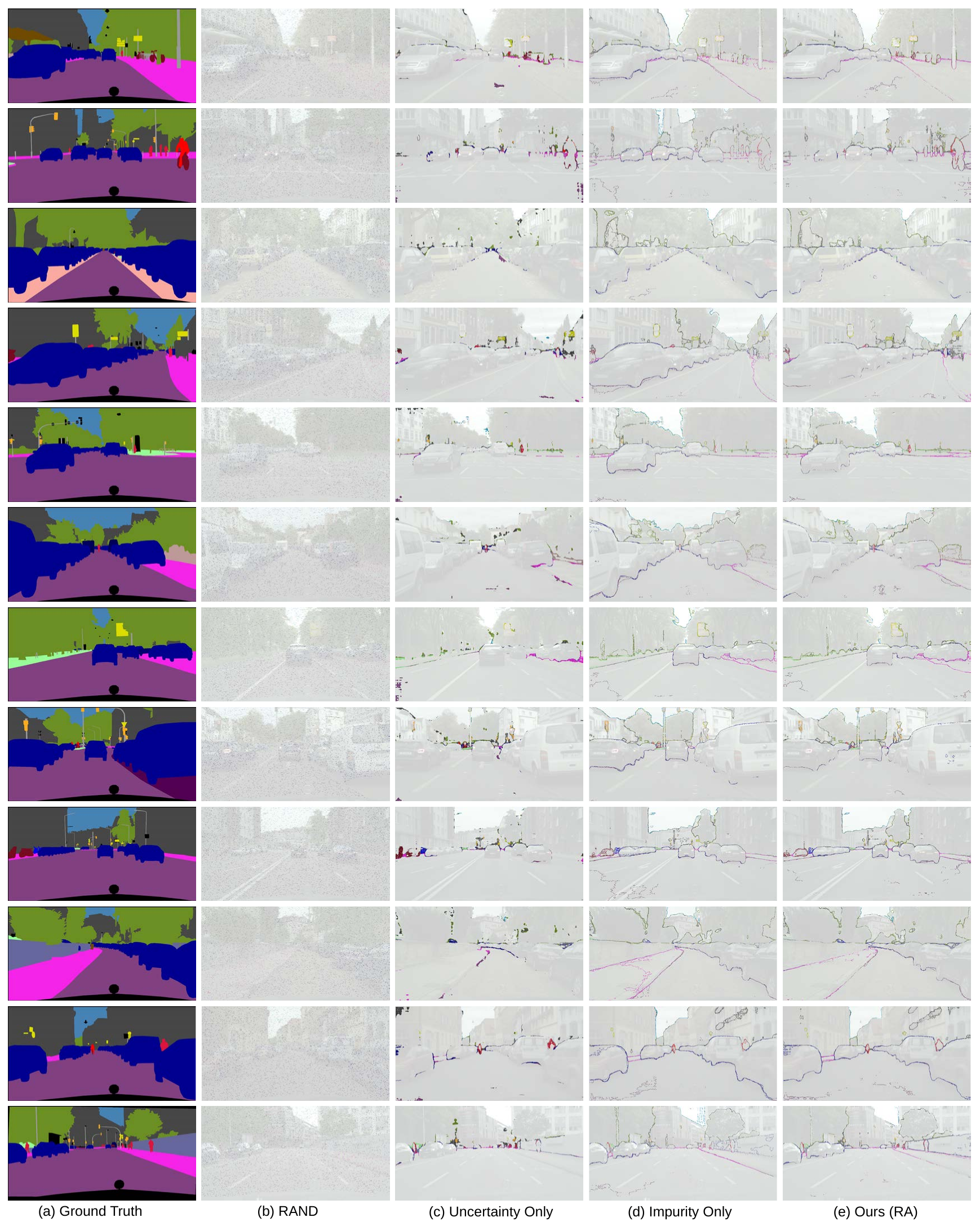} \vspace{-1.5mm}
    \caption{
      {\bf Visualization of queried regions to annotate (2.2\%) on GTAV $\to$ Cityscapes.} Compared to RAND, Uncertainty Only, and Impurity Only, Ours (\region) is able to select the most diverse and uncertain regions of an image. Please zoom in to see the details.
    } \vspace{-4mm}
    \label{fig:selected_gta}
\end{figure*}

\begin{figure*}
  \centering  
    \includegraphics[width=0.98\textwidth]{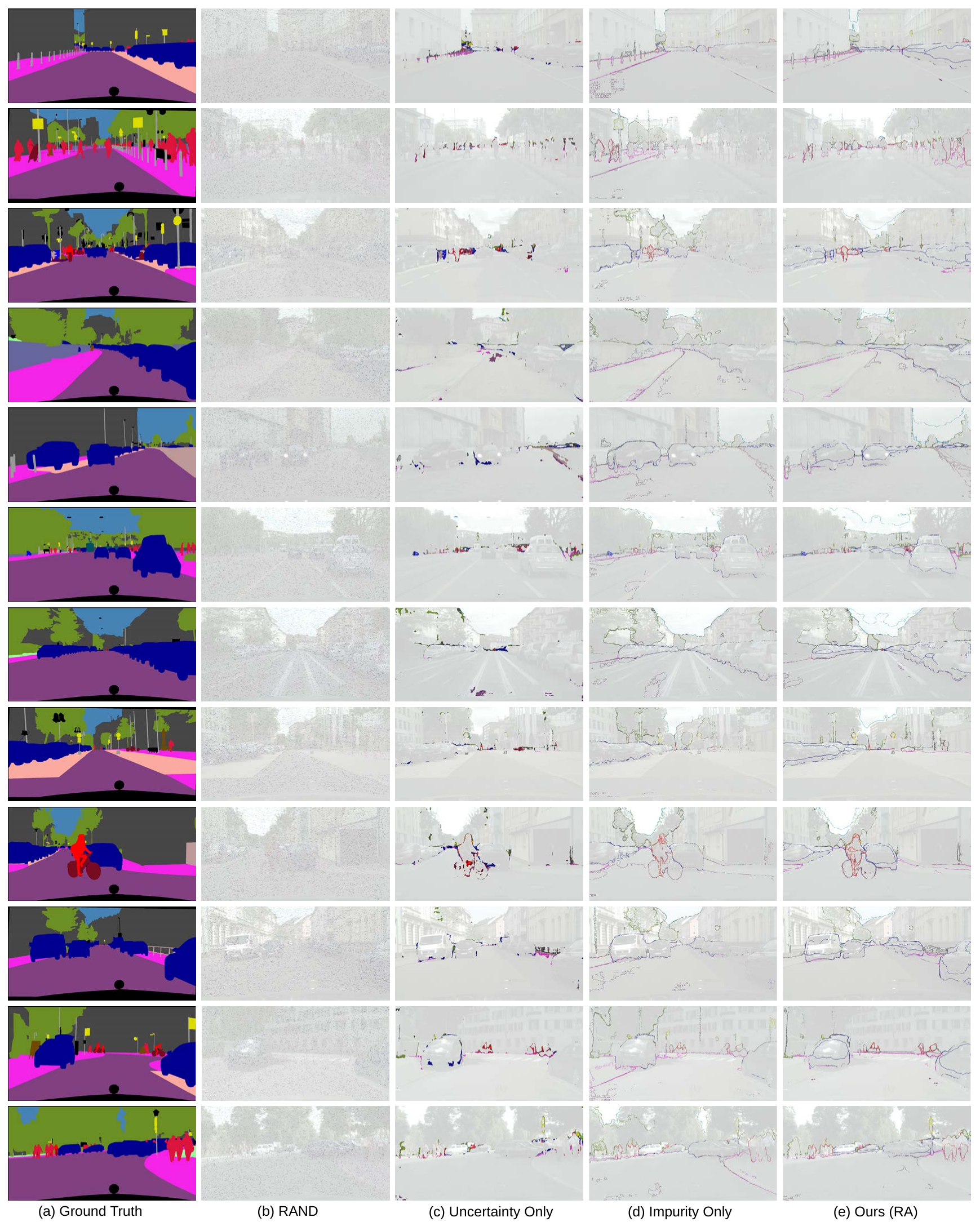} \vspace{-1.5mm}
    \caption{
      {\bf Visualization of queried regions to annotate (2.2\%) on SYNTHIA $\to$ Cityscapes.} Compared to RAND, Uncertainty Only, and Impurity Only, Ours (\region) is able to select the most diverse and uncertain regions of an image. Please zoom in to see the details.
    } \vspace{-4mm}
    \label{fig:selected_syn}
\end{figure*}

We follow the same conventions as Fig.~\ref{fig:visulization_results} and Fig.~\ref{fig:visulization_selected} of the main paper, and present additional results for qualitative comparisons under various settings, including GTAV $\to$ Cityscapes (Fig.~\ref{fig:results_gta} and Fig.~\ref{fig:selected_gta}), SYNTHIA $\to$ Cityscapes (Fig.~\ref{fig:results_syn} and Fig.~\ref{fig:selected_syn}).

\end{document}